\documentclass[nohyperref]{article}

\usepackage{microtype}
\usepackage{subfigure}

\usepackage{url}
\usepackage{bbm}
\usepackage{bm}
\usepackage{float}
\usepackage{graphicx}
\usepackage{amsfonts}
\usepackage{booktabs}
\usepackage{xcolor}
\usepackage{multirow}
\usepackage{multicol}
\usepackage{makecell}

\usepackage{hyperref}

\usepackage[accepted]{icml2023}

\usepackage{amsmath}
\usepackage{amssymb}
\usepackage{mathtools}
\usepackage{amsthm}

\usepackage[capitalize,noabbrev]{cleveref}

\theoremstyle{plain}
\newtheorem{theorem}{Theorem}[section]
\newtheorem{proposition}[theorem]{Proposition}
\newtheorem{lemma}[theorem]{Lemma}

\theoremstyle{definition}

\theoremstyle{remark}

\newenvironment{sizeddisplay}[1]
 {\par\nopagebreak#1\noindent\ignorespaces}
 {\nopagebreak\ignorespacesafterend}

\usepackage[textsize=tiny]{todonotes}

\icmltitlerunning{Bayesian Neural Networks Avoid Encoding Complex and Perturbation-Sensitive Concepts}

\begin{document}

\twocolumn[
\icmltitle{Bayesian Neural Networks Avoid Encoding  Complex and \\ Perturbation-Sensitive Concepts}



\icmlsetsymbol{equal}{*}

\begin{icmlauthorlist}
\icmlauthor{Qihan Ren}{equal,school}
\icmlauthor{Huiqi Deng}{equal,school}
\icmlauthor{Yunuo Chen}{school}
\icmlauthor{Siyu Lou}{school}
\icmlauthor{Quanshi Zhang}{school,note}
\end{icmlauthorlist}

\icmlaffiliation{school}{Shanghai Jiao Tong University, Shanghai, China}
\icmlaffiliation{note}{Quanshi Zhang is the corresponding author. He is with the Department of Computer Science and Engineering, the John Hopcroft Center, at the Shanghai Jiao Tong University, China}

\icmlcorrespondingauthor{Quanshi Zhang}{zqs1022@sjtu.edu.cn} % needs revision

\icmlkeywords{Machine Learning, ICML}

\vskip 0.3in
]



\printAffiliationsAndNotice{\icmlEqualContribution} % otherwise use the standard text.

\begin{abstract}
In this paper, we focus on mean-field variational Bayesian Neural Networks (BNNs) and explore the representation capacity of such BNNs by investigating which types of concepts are less likely to be encoded by the BNN.
It has been observed and studied that a relatively small set of interactive concepts usually emerge in the knowledge representation of a sufficiently-trained neural network, and such concepts can faithfully explain the network output. Based on this, our study proves that compared to standard deep neural networks (DNNs), it is less likely for BNNs to encode complex concepts.
Experiments verify our theoretical proofs. Note that the tendency to encode less complex concepts does not necessarily imply weak representation power, considering that complex concepts exhibit low generalization power and high adversarial vulnerability. The code is available at \url{https://github.com/sjtu-xai-lab/BNN-concepts}.
\end{abstract}

%%%%%%%%% BODY TEXT
\section{Introduction}
Unlike standard deep neural networks (DNNs), Bayesian neural networks (BNNs) represent network weights as probability distributions. Therefore, BNNs exhibit distinctive representation capacities from standard DNNs. Existing studies~\cite{blundell2015weight,gal2018sufficient,kristiadi2020being,carbone2020robustness, wenzel2020good, krishnan2020specifying,zhang2022improving} usually analyzed BNNs in terms of generalization power, adversarial robustness, and optimization.

In contrast to the above studies, this paper proposes a new perspective to investigate the representation capacity of BNNs, \textit{i.e.}, we discover and theoretically prove that BNNs are less likely to encode complex and perturbation-sensitive concepts than standard DNNs. In fact, such a property brings specific advantages to feature representations of BNNs.
To be precise, we limit our study to the scope of \textbf{mean-field variational BNNs}~\cite{blundell2015weight}, which is one of the most commonly used BNNs. Thus, in this paper, we just use the term \textit{BNN} to refer to mean-field variational BNNs.

\textbf{Common phenomenon of concept emergence in various neural networks.} 
Although it is well-known that a neural network does not explicitly encode concepts like graphical models, recent studies have discovered~\cite{ren2021AOG,li2023does} and theoretically proved~\cite{ren2023where} a common concept-emerging phenomenon that neural networks usually \textit{implicitly} encode a small number of interactive concepts for inference, which have been observed
in different neural networks for various tasks. Specifically,  each interactive concept represents an AND relationship among a set of input variables.

For example, we can use {\small $I(S=\{\text{eyes, nose, mouth}\}) = U_S\cdot\mathrm{exist(\text{eyes})}\cdot\mathrm{exist(\text{nose})}\cdot\mathrm{exist(\text{mouth})}$} to illustrate the AND relationship for the face concept  in image classification. If any image patch in the set {\small $S = \{\text{eyes, nose, mouth}\}$}  is masked, then the face concept will be deactivated, and the numerical effect of this concept is removed ({\small $I(S)=0$}) and no longer influences the network output.

\begin{figure*}\centering
\includegraphics[ width=0.88\textwidth]{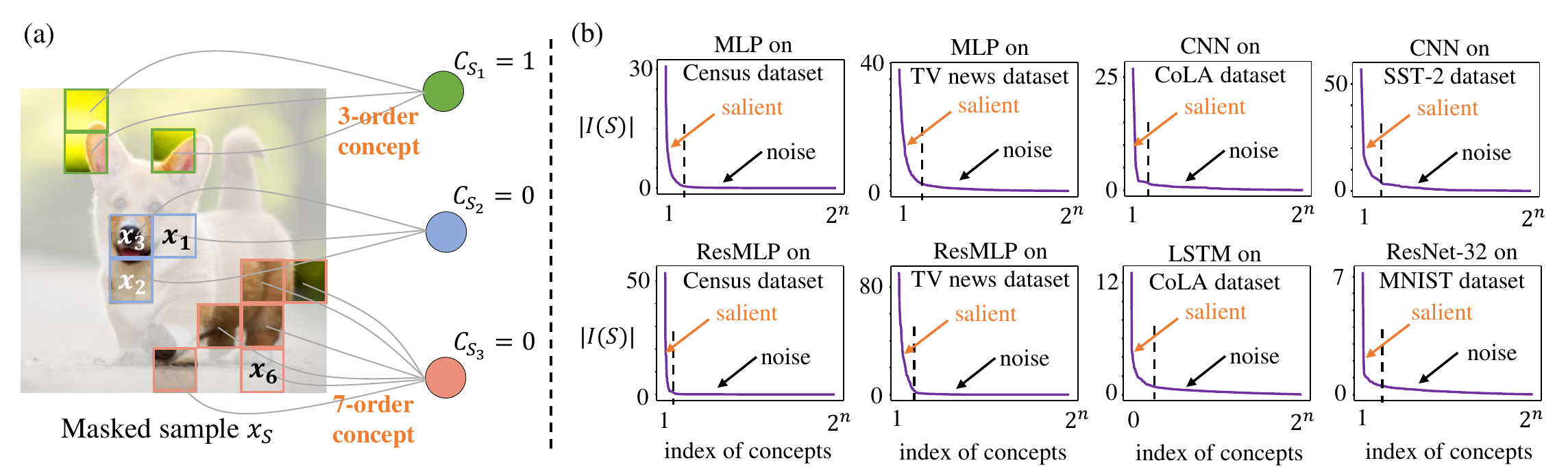}
\vspace{-0.15cm}
\caption{(a) Illustration of interactive concepts encoded by a neural network. Each interactive concept {\small $S$} corresponds to an AND relationship among a specific set {\small $S$} of input variables (image patches). {\small$C_S$} represents the activation state of the concept {\small $S$}. 
The patch {\small $x_1$} is masked ({\small ${\rm unmask}(x_1)=0$}), so the concept {\small $S_2$} is deactivated, \emph{i.e.}, {\small$C_{S_2}=\bigwedge_{i=1}^3 {\rm unmask}(x_i)=0$}. 
Similarly, {\small $x_6$} is masked, so that the concept {\small $S_3$} is deactivated, and {\small$C_{S_3} =0$}.
(b) Experiments demonstrate the common concept-emerging phenomenon. Neural networks with various architectures all encode sparse interactive concepts. In other words, most interactive concepts have near-zero effects, \textit{i.e.}, {\small $I(S)\approx 0$}, and can be considered as noises; only a relatively small number of interactive concepts have significant effects. For better visualization, the interactive concepts are sorted by strengths in descending order.}
\label{fig:causal-graph}
\vspace{-0.15cm} % todo
\end{figure*}

More importantly,  interactive concepts can be considered as faithful inference patterns encoded by the neural network. It is because~\citet{ren2021AOG} has proved that people can use a relatively small number of interactive concepts to well mimic the inference logic of the neural network on a certain input sample. That is, numerical effects of these concepts always well predict diverse network outputs, no matter how the input sample is masked.

\textbf{BNNs ignore complex and perturbation-sensitive concepts.} 
Based on the interactive concepts, we discover and theoretically prove that \textit{compared to standard DNNs, it is more difficult for a neural network to encode complex interactive concepts, as long as it has weight uncertainty.}
The complexity of an interactive concept {\small $S$} is defined as the number of variables in the set {\small $S$}, \textit{i.e.}, {\small ${\rm complexity}(S) = |S|$}. 
Here, {\small $|S|$} is also termed the \textit{order} of the interactive concept.

We prove the above conclusion  through three steps. First, it is difficult to theoretically analyze interactive concepts encoded by BNNs, because BNNs represent network weights as probability distributions.
To this end, we find that we can usually use a \textit{surrogate DNN model}, which is constructed by adding perturbations to both the input and low-layer features of a standard DNN, to approximate feature representations of a BNN.
In this way, we can directly analyze the surrogate DNN model with feature uncertainty, instead of investigating the BNN with weight uncertainty.

Second,  we prove that in the surrogate DNN model, high-order interactive concepts are more sensitive to random perturbations than low-order interactive concepts.

Third, we prove that the sensitivity makes high-order interactive concepts difficult to be learned when features are perturbed.
In this way, we can conclude that high-order interactive concepts are also less likely to be learned by the BNN when its weights are perturbed.

In addition, experiments showed that the strength of high-order (complex) interactive concepts encoded by BNNs was weaker than those encoded by standard DNNs, which verified the above theoretical conclusion.

\textbf{Note that our proof does {NOT} mean that a BNN has limited representation capacity.} Instead, we just demonstrate the distinctive tendency of avoiding encoding complex (high-order) interactive concepts, when weight uncertainty is introduced into the neural network. This does not mean that BNNs have weaker representation power than standard DNNs. If the task loss requires to encode complex concepts, then our research indicates that the BNN must reduce its weight uncertainty, to some extent.

\textbf{Practical values and advantages of avoiding encoding complex concepts.} Although we prove that BNNs tend to avoid encoding complex concepts, it is not necessarily a disadvantage of the BNN, compared to standard DNNs.
On the contrary, it has been found that compared to simple (low-order) concepts, complex (high-order) concepts encoded by a neural network usually have poorer generalization ability~\cite{pmlr-v151-lengerich22a} and are more vulnerable to adversarial attacks~\cite{ren2021towards}.
Thus, encoding less complex concepts might be an advantage.

\section{BNNs ignore complex and perturbation- sensitive concepts}
Unlike standard DNNs, a BNN represents each weight in the network as a probability distribution, instead of a scalar. In this paper, we limit the scope of our study to mean-field variational BNNs~\cite{blundell2015weight}, where all weights {\small $\boldsymbol{W}$} are formulated as a Gaussian distribution {\small $\mathcal{N}(\boldsymbol{W};\boldsymbol{\mu},\boldsymbol{\Sigma})$}, 
and the covariance matrix {\small $\boldsymbol{\Sigma}$} is diagonal. Other types of BNNs (\emph{e.g.}, BNNs based on the Monte Carlo Dropout~\cite{gal2016dropout}) are not discussed. The BNN learns parameters 
{\small $\boldsymbol{\theta}=(\boldsymbol{\mu}, \boldsymbol{\Sigma})$}, and we use {\small $q_{\boldsymbol{\theta}}(\boldsymbol{W})$} to represent the weight distribution.
Let us consider a classification task with the training data {\small $\mathcal{D}=$ $\{(\boldsymbol{x}^{(1)},y^{(1)}),\dots,(\boldsymbol{x}^{(n)},y^{(n)})\}$}. 
Training a BNN is to minimize the Kullback-Leibler (KL) divergence between the distribution {\small $q_{\boldsymbol{\theta}}(\boldsymbol{W})$} and the posterior distribution {\small $p(\boldsymbol{W}|\mathcal{D})$}.
\begin{sizeddisplay}{\small}
% \begin{small}
\begin{align} \label{eq:BNN-training}
% \vspace{-0.2cm}
\boldsymbol{\theta}^{*}\!\! &=\arg \! \min_{\boldsymbol{\theta}} {\rm KL}[q_{\boldsymbol{\theta}}(\boldsymbol{W}) \Vert p(\boldsymbol{W} | \mathcal{D})] \\
&=\arg \! \min_{\boldsymbol{\theta}} - \mathbb{E}_{\boldsymbol{W} \sim q_{\boldsymbol{\theta}}(\boldsymbol{W})}[\log p(\mathcal{D} | \boldsymbol{W})] 
\!+\! {\rm KL}[q_{\boldsymbol{\theta}}(\boldsymbol{W}) \Vert p(\boldsymbol{W})], \nonumber
% \vspace{-0.2cm}
\end{align}
% \end{small}%
\end{sizeddisplay}%
where the first term is the classification loss, and the second term is the KL divergence between {\small $q_{\boldsymbol{\theta}}(\boldsymbol{W})$} and the prior distribution {\small $p(\boldsymbol{W})$}, which is usually formulated as a Gaussian distribution {\small $\mathcal{N}(\boldsymbol{W};\boldsymbol{0},\boldsymbol{I})$}.
In addition, given a testing sample {\small $\boldsymbol{x}$}, the inference of the BNN is conducted as follows. First, network weights are sampled from the weight distribution {\small $q_{\boldsymbol{\theta}}(\boldsymbol{W})$} to construct multiple neural networks. Then, each network is used to conduct inference on the sample {\small $\boldsymbol{x}$}, and the final inference result $p(y|\boldsymbol{x})$ is computed as the average classification probability of all the networks,
\begin{equation}\label{eq:BNN-inference}
     p(y | \boldsymbol{x}) = \mathbb{E}_{\boldsymbol{W} \sim q_{\boldsymbol{\theta}}(\boldsymbol{W})}[p(y|\boldsymbol{x}, \boldsymbol{W})]. 
\end{equation}

\subsection{Preliminaries: emergence of sparse concepts}
\label{sec:representing}

The learning of neural networks is usually regarded as a fitting problem between the ground-truth label and the model prediction, without explicit learning of specific concepts.
However,  recent studies
have empirically discovered~\cite{ren2021AOG,li2023does} and theoretically proved~\cite{ren2023where} that sparse AND relationships between input variables were usually implicitly encoded by a neural network  when it was sufficiently trained.
As shown in Figure \ref{fig:causal-graph}(a), these AND relationships can be viewed as specific types of \textit{interactive concepts}, which will be introduced in the \textbf{interactive concepts} paragraph.

Although counter-intuitive, this concept-emerging phenomenon does exist in various neural networks. 
Furthermore, such interactive concepts have been used to prove the representation bottleneck of the neural network \cite{deng2022discovering} and obtain optimal masking states for attribution methods~\cite{ren2023can}.
\textbf{We also verify the trustworthiness of using interactive concepts to explain neural networks in experiments (see the end of this section).}

\textbf{Interactive concepts.} ~\citet{ren2021towards} proposed the interaction effect {\small $I(S)$} to study the emergence of concepts.
Let us consider a pre-trained neural network $v$ and an input sample {\small$\bm{x} = [x_1, \dots, x_n]$} with $n$ input variables  indexed by $N=\{1, \dots, n\}$.
Let {\small$\Omega$} denote a set of interactive concepts extracted from the network. Each interactive concept {\small$S\in\Omega$} corresponds to the collaboration (AND relationship) between input variables in a specific set {\small $S\subseteq N$}, thus {\small $\Omega \subseteq 2^N =\{S|S\subseteq N\}$}.
For instance, as Figure \ref{fig:causal-graph}(a) shows,  a concept {\small $S=\{x_1, x_2, x_3\}$} is formed due to the co-occurrence of the three image patches.
The concept will be activated and make a certain interaction effect {\small $I(S)$} on the network output, only if the patches {\small $x_1, x_2, x_3$} are all present.
In contrast, the absence (masking) of any patch among {\small $x_1$}, {\small $x_2$}, and {\small $x_3$} will deactivate the concept and remove the interaction effect, \textit{i.e.}, {\small $I(S|\bm{x}^{\text{mask}}) = 0$}.

Specifically, the interaction effect {\small $I(S|\bm{x})$} on the sample {\small $\bm{x}$} is computed by the Harsanyi dividend~\cite{harsanyi1963}.
\begin{equation}\label{eqn:harsanyi}
	\begin{small}
		\begin{aligned}
		I(S|\bm{x}) = {\sum}_{T \subseteq S}(-1)^{|S|-|T|}\cdot v(\bm{x}_{T}).
	  \end{aligned}
	\end{small}
\end{equation}
If {\small $I(S|\bm{x})$} has a significant value, then the neural network is considered to encode an interactive concept {\small $S$}; otherwise, if {\small $I(S|\bm{x}) \approx 0$}, the concept {\small $S$} does not exist.
Here, {\small$\bm{x}_T$} denotes the masked input sample, where variables in {\small$N\setminus T$} are masked and variables in {\small$T$} are kept unchanged.
Besides, {\small$v(\bm{x}_T) \in \mathbb{R}$} can be computed as a scalar output of the neural network on the masked sample {\small $\bm{x}_T$} (\textit{e.g.}, the confidence score of classifying the input sample {\small$\bm{x}_T$} to the ground-truth category {\small$v(\bm{x}_T) = \text{log} \frac{p(y=y_\text{truth}|\bm{x}_T)}{1-p(y=y_\text{truth}|\bm{x}_T)}$}).

\textbf{Faithfulness of interactive concepts.} 
Given an input sample {\small$\bm{x}$} with {\small $n$} variables, we have {\small$2^{n}$} different ways to mask the sample {\small$\bm{x}$} and obtain the masked sample {\small$\bm{x}_T$} \textit{w.r.t.} all subsets {\small$T \subseteq N$}.
To this end, \citet{ren2021towards} proved that
\begin{equation}
	\begin{small}
		\begin{aligned}
		\label{eqn:effciency}
		\exists \ \Omega \subseteq 2^N, \  s.t. \  \forall \ T \subseteq N, v(\bm{x}_T) = \sum\nolimits_{S \in \Omega, S \subseteq T} I(S|\bm{x}),
			  \end{aligned}
	\end{small}
\end{equation}
where {\small $2^N = \{S | S\subseteq N\}$}. The equation indicates that interactive concepts in  {\small $\Omega$} can well mimic network outputs on all the {\small $2^n$} masked samples. Thus, we can consider that all interactive concepts in the set {\small $\Omega$} as faithful inference patterns encoded by the 
neural network.

\textbf{Sparsity of interactive concepts.} More crucially, extensive experiments~\cite{ren2021AOG,li2023does} discovered that interactive concepts emerging in a neural network are usually very sparse. 
Figure \ref{fig:causal-graph}(b) shows that most interactive concepts have near-zero interaction effects ({\small$|I(S|\bm{x})|\approx 0$}), thus having negligible influence on the network output.
Only a few salient interactive concepts have significant effects {\small$I(S|\bm{x})$} on the network output. In this way, the network output can be mimicked by only a few salient interactive concepts in {\small $\Omega_{\text{salient}}$}.
\begin{equation}
	\begin{small}
		\begin{aligned}
		\label{eqn:salient_residual}
		  \forall \ T \subseteq N, \ \ v(\bm{x}_T) &= \sum\nolimits_{S \in \Omega_{\text{salient}}} I(S|\bm{x}) + \epsilon
		\end{aligned}
	\end{small}
\end{equation}
The above equation decomposes the output {\small$v(\bm{x}_T)$} into two parts: (1) effects of all salient interactive concepts in {\small $\Omega_{\text{salient}}$}, and (2) a small residual term $\epsilon$ containing negligible effects of all non-salient interactive concepts. 

\textbf{Empirically verifying the sparsity of concepts.} 
Based on Eq.~(\ref{eqn:salient_residual}), in the following analysis, \textit{only salient interactive concepts  in {\small $\Omega_{\text{salient}}$} are regarded as valid concepts encoded by a neural network.} We empirically verify the emergence of sparse concepts in various neural networks, including
multi-layer perceptrons (MLPs), residual multi-layer perceptrons (ResMLPs)~\cite{resmlp}, long short-term memory (LSTM)~\cite{lstm}, and
convolutional neural networks (CNNs), and on different datasets, including tabular data (Census dataset and TV news dataset~\cite{UCIrepository}), language data (CoLA~\cite{warstadt2019neural} and SST-2~\cite{socher2013recursive}), and image data (MNIST~\cite{lecun1998gradient}). Figure \ref{fig:causal-graph}(b) verifies that  concepts encoded by various neural networks are all sparse.

\textbf{Complexity of a neural network representing a concept}.
In many previous studies~\cite{deng2022discovering, wang2021unified, zhang2020interpreting}, the complexity of an
interactive concept {\small $S$} was measured by the number of variables in the set {\small $S$} (also termed the \textit{order} of the interactive concept), \emph{i.e.}, {\small ${\rm complexity}(S)={\rm order}(S)=|S|$}.
Then,  a low-order concept represents a simple collaboration among a few input variables, 
while a high-order concept represents a complex collaboration among many input variables.

\subsection{Approximating weight uncertainty by adding input perturbations}\label{sec:approximation}
In this paper, we aim to prove that compared to standard DNNs, it is more difficult to encode high-order (complex) interactive concepts as long as the network has weight uncertainty.
Note that previous studies~\cite{pmlr-v151-lengerich22a, ren2021towards} found that a DNN encoding less complex concepts was \textbf{NOT} necessarily equivalent to a weak representation capacity. Instead, it usually boosts the generalization power and adversarial robustness. 
In addition, as discussed in the last two paragraphs of the introduction, the BNN can still encode complex concepts when it learns small variances.

Unlike standard DNNs, a BNN formulates each weight as a probability distribution, which boosts the difficulty of theoretically analyzing interactive concepts encoded in a BNN. 
Therefore, in this subsection, we first discover that introducing uncertainty to weights in the BNN can be approximated by adding perturbations to input variables and low-layer features in experiments. In other words, we add random perturbations to both input variables and low-layer features of a standard DNN, and we demonstrate that such a perturbed DNN performs as a \textit{surrogate DNN model}, which well approximates feature representations of a BNN.

Let us consider a feed-forward BNN, which has $L$ cascaded linear layers and ReLU layers. Given an input sample {\small $\boldsymbol{x}
\in \mathbb{R}^{D_0}$} ({\small $D_0 = n$}), the feature of the $l$-th layer {\small$\boldsymbol{h}^{(l)}\in \mathbb{R}^{D_l}$ ($1\le l \le L$)} is computed as follows.
\begin{equation}
\begin{small}
\begin{aligned}
\boldsymbol{h}^{(l)} = \boldsymbol{W}^{(l)}(\cdots \boldsymbol{\Phi}^{(1)} (\boldsymbol{W}^{(1)} \boldsymbol{x} + \boldsymbol{b}^{(1)})\cdots)+\boldsymbol{b}^{(l)}, 
\end{aligned}
\end{small}
\end{equation}
where {\small $\boldsymbol{W}^{(l)}\in \mathbb{R}^{D_l \times D_{l-1}}$} and  {\small $\boldsymbol{b}^{(l)}\in \mathbb{R}^{D_l}$} 
denote the weight matrix and bias of the $l$-th linear layer, respectively. 
In the BNN, {\small $W_{ij}^{(l)} \sim \mathcal{N}(\overline W_{ij}^{(l)}, (\sigma_{ij}^{(l)})^{2})$} is independently sampled from Gaussian distributions. 
We use {\small $\boldsymbol{\mu}_{\boldsymbol{W}^{(l)}}=[\overline W_{ij}^{(l)}] \in \mathbb{R}^{D_l \times D_{l-1}}$} to denote
the mean of the weight matrix.
Besides, {\small $\boldsymbol{b}^{(l)} \sim \mathcal{N}(\boldsymbol{\mu}_{\boldsymbol{b}^{(l)}},\boldsymbol{\Sigma}_{\boldsymbol{b}^{(l)}})$}, where 
{\small $\boldsymbol{\Sigma}_{\boldsymbol{b}^{(l)}}$} is a diagonal matrix.
The diagonal matrix {\small $\boldsymbol{\Phi}^{(l)}={\rm diag}(\phi_1^{(l)}, \cdots, \phi_{D_l}^{(l)})\in \{0,1\}^{{D_l}\times {D_l}}$} denotes binary gating states of the $l$-th ReLU layer.

\begin{figure}[t]
\centering
\includegraphics[ width=0.96\linewidth]{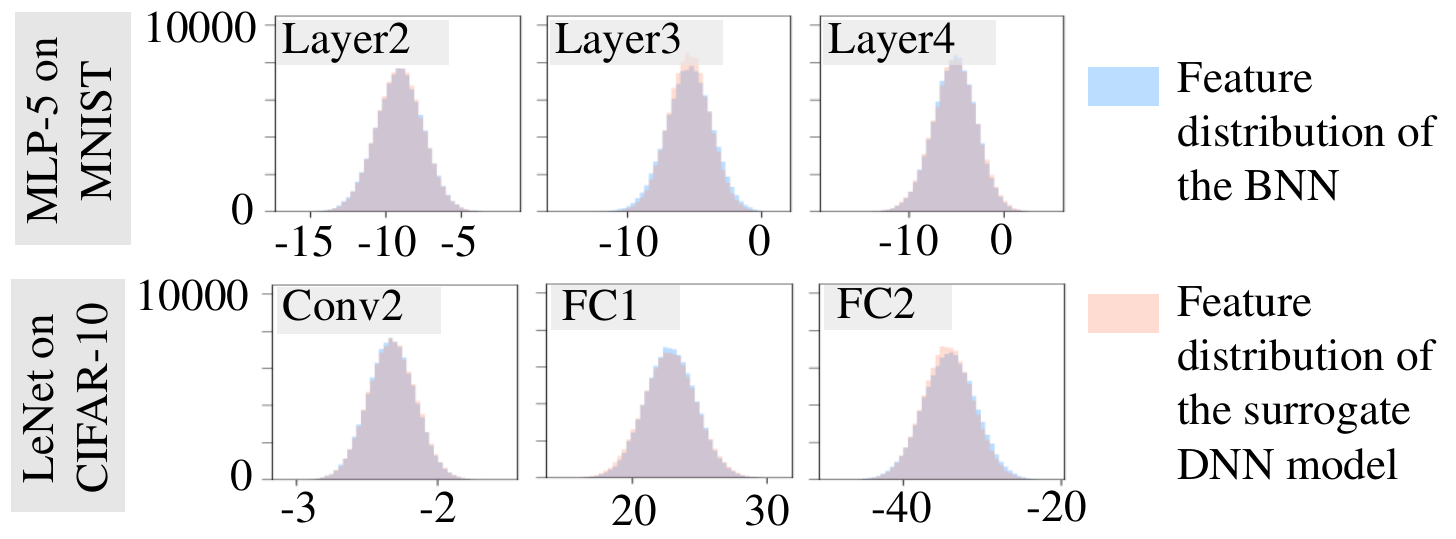}
% \vspace{-0.27cm}
\caption{Comparison between the feature distribution of the BNN and the feature distribution of the surrogate DNN model. 
We randomly selected a feature dimension from each layer of the network. Each sub-figure compares feature distributions between the BNN and the surrogate DNN model in the selected dimension.
Please see Appendix \ref{sec:apdx_visual_result} for results on tabular datasets.}
% \vspace{-0.35cm}
\label{fig:step1-exp}
\end{figure}

\begin{table}[t]
\centering
    \caption{Approximation error of the surrogate model and approximation error of the baseline distribution. The approximation error was measured using features of the last layer of the network.}
    \vspace{0.05cm}
\begin{footnotesize}
\begin{tabular}{lcccc} 
\toprule
 \multirow{2}*{}\!\!\! & {\footnotesize MLP-5 on}\!\!\!  & {\footnotesize LeNet on}\!\!\! & {\footnotesize MLP-8 on}\!\!\! & {\footnotesize MLP-8 on}  \\
& {\footnotesize MNIST}\!\!\! & {\footnotesize CIFAR-10}\!\!\! & {\footnotesize Census}\!\!\! & {\footnotesize TV news}  \\ \hline
{\small surrogate} & \textbf{0.16} &  \textbf{0.06}  & \textbf{0.11} & \textbf{0.16}\\
{\small baseline} & 21.38 & 19.68 & 4.79 & 4.50\\ \bottomrule
  \end{tabular}
  % \vspace{-0.35cm}
  \label{tab:approx_error}
  \end{footnotesize}
\end{table}

Then, we construct the surrogate DNN model with the same architecture as the BNN, to approximate the BNN's feature distribution. Parameters of this surrogate DNN model {\small $\boldsymbol{\psi}$} are set as the mean of the weight distribution and the mean of the bias distribution in the BNN, 
 \emph{i.e.}, {\small $\boldsymbol{\psi}=\{\boldsymbol{\mu}_{\boldsymbol{W}^{(l)}}, \boldsymbol{\mu}_{\boldsymbol{b}^{(l)}}\}_{l=1}^L$}. 
Given an input sample {\small $\bm{x}$}, we add perturbations {\small $\Delta \boldsymbol{x} \sim \mathcal{N}(\boldsymbol{0}, \boldsymbol{\Sigma}_{\Delta \boldsymbol{x}})$} to input variables and perturbations {\small $\Delta \boldsymbol{h}^{(l')} \sim \mathcal{N}(\boldsymbol{0}, \boldsymbol{\Sigma}_{\Delta \boldsymbol{h}^{(l')}})$}
to features between the first layer and the $(l-1)$-th layer in the surrogate DNN model ({\small $1\le l' \le l-1$}). 
In this way, we can obtain the distribution of the $l$-th layer feature  $\tilde{h}^{(l)}$ in the surrogate DNN model, 
denoted as {\small $p_{\text{DNN}}(\boldsymbol{\tilde{h}}^{(l)}|\boldsymbol{\Delta}= \{\bm{\Sigma}_{\Delta \bm{x}}, \bm{\Sigma}_{\Delta \bm{h}^{(1)}}, \cdots, \bm{\Sigma}_{\Delta \bm{h}^{(l-1)}}\})$}, 
and we use {\small $p_{\text{DNN}}(\boldsymbol{\tilde{h}}^{(l)}|\boldsymbol{\Delta})$} to mimic the feature distribution {\small $p_{\text{BNN}}(\boldsymbol{h}^{(l)})$} in the BNN.    
Thus, the objective function is formulated as minimizing the following KL divergence.
\begin{equation}\label{eqn:mimicBNN}
\begin{small}
  \ \forall \ 1 \leq l \leq L,  \quad \min_{\boldsymbol{\Delta}}
   {\rm KL}(p_{\text{BNN}}(\boldsymbol{h}^{(l)}) \Vert  p_{\text{DNN}}(\boldsymbol{\tilde{h}}^{(l)}|\boldsymbol{\Delta})), 
 \end{small}
\end{equation}
where we set {\small $\boldsymbol{\Sigma}_{\Delta \boldsymbol{x}}, \boldsymbol{\Sigma}_{\Delta \boldsymbol{h}^{(l')}} \in \bm{\Delta}$} as diagonal matrices.

However, it is difficult to directly optimize Eq.~(\ref{eqn:mimicBNN}). 
Instead, we learn the covariance matrices 
in a layer-wise manner, as follows.
First, we learn the covariance matrix {\small  $\boldsymbol{\Sigma}_{\Delta \boldsymbol{x}}$} on input variables to match the first-layer feature of the surrogate DNN model to the first-layer feature of the BNN, \textit{i.e.}, 
{\small $\min_{\boldsymbol{\Sigma}_{\Delta \boldsymbol{x}}} $ ${\rm KL}(p_{\text{BNN}}(\boldsymbol{h}^{(1)}) \Vert p_{\text{DNN}}(\boldsymbol{\tilde{h}}^{(1)}| \boldsymbol{\Sigma}_{\Delta \boldsymbol{x}}))$}.
Then, we fix the learned covariance matrix {\small $\boldsymbol{\Sigma}_{\Delta \boldsymbol{x}}$} (note that it is not to fix the perturbation {\small $\Delta \boldsymbol{x}$}), and learn the covariance matrix {\small $\boldsymbol{\Sigma}_{\Delta \boldsymbol{h}^{(1)}}$} on the first-layer feature to fit feature distributions of the second layer
by  minimizing  {\small ${\rm KL}(p_{\text{BNN}}(\boldsymbol{h}^{(2)}) \Vert  p_{\text{DNN}}(\boldsymbol{\tilde{h}}^{(2)}|\boldsymbol{\Sigma}_{\Delta \boldsymbol{x}}, \boldsymbol{\Sigma}_{\Delta \boldsymbol{h}^{(1)}}))$}.
We recursively learn the covariance matrix of an upper layer by fixing the covariance matrices in all lower layers, until the last layer.

\textbf{Experimental verification.}
We trained BNNs on image datasets and tabular datasets to verify the quality of using the surrogate DNN model to approximate the feature distribution of the BNN.
For image datasets, we tested BNNs with two architectures. For the MNIST dataset~\cite{lecun1998gradient}, we constructed  a BNN with the architecture of a 5-layer MLP. We also tested a BNN with the LeNet architecture~\cite{lecun1998gradient}, which was trained on the CIFAR-10 dataset~\cite{krizhevsky2009learning}.
We used two tabular datasets, including the UCI TV news dataset (termed \textit{TV news}) and the UCI census income  dataset (termed \textit{Census})~\cite{UCIrepository}. We constructed BNNs with an 8-layer MLP architecture for these tabular datasets.
All MLPs contained 100 neurons in each hidden layer. 	
For each BNN, we constructed a corresponding surrogate DNN model. Please see Appendix \ref{sec:apdx_exp_detail} for implementation details.

Figure \ref{fig:step1-exp} shows that the feature distribution of the surrogate DNN model well matched the feature distribution of the BNN.  
Furthermore, we used the KL divergence {\small ${\rm KL}(p_{\text{BNN}}(\bm{h}^{(l)}) \Vert  p_{\text{DNN}}(\bm{\tilde{h}}^{(l)}|\bm{\Delta}))$} in Eq.~(\ref{eqn:mimicBNN}) to measure the approximation error. 
To compare with {\small ${\rm KL}(p_{\text{BNN}}(\bm{h}^{(l)}) \Vert  p_{\text{DNN}}(\bm{\tilde{h}}^{(l)}|\bm{\Delta}))$}, we further constructed a simple baseline distribution of the features {\small $p_{\text{base}}(\bm{h}^{(l)})=\mathcal{N}(\hat{\mu}\bm{1}, \hat{\sigma}^2\bm{I})$}, 
where {\small $\hat{\mu}$} and {\small $\hat{\sigma}^2$} denote the mean and the variance over all feature dimensions of the BNN, respectively. 
We computed {\small ${\rm KL}(p_{\text{BNN}}(\bm{h}^{(l)}) \Vert p_{\text{base}}(\bm{h}^{(l)}))$} for comparison.
Table \ref{tab:approx_error} shows that the approximation error of the surrogate DNN model was significantly smaller than the approximation error of the baseline distribution.

Experimental results showed that the weight uncertainty in a BNN could be well approximated by adding random perturbations to both input variables and low-layer features.

\subsection{High-order concepts are sensitive to perturbations}
\label{sec:sensitivity}
In the previous subsection, we have demonstrated that adding random perturbations to input variables and low-layer features can successfully approximate the feature distribution in a BNN with weight uncertainty. In this way, proving the difficulty of BNNs in encoding high-order interactive concepts can be converted into the proof of the following two steps.
First, in this subsection, we prove that high-order interactive concepts are more sensitive to perturbations than low-order interactive concepts, which is inspired by the proof in~\citet{zhou2023concept}. Then, in the next subsection, we will prove that perturbation-sensitive concepts are difficult to be learned by a neural network.

Note that according to Section \ref{sec:approximation}, introducing the weight uncertainty in a BNN can be approximated by adding random perturbations to both input variables and features of different layers. However, simultaneously adding perturbations to features of multiple layers significantly boosts the difficulty of analysis. 
Fortunately, adding perturbations to output features of the $l$-th layer can be considered as perturbing input variables of the $(l+1)$-th layer. Hence, in this subsection, we just analyze interactive concepts in a simple case where we  perturb input variables in a certain layer, instead of analyzing the complex case of simultaneously perturbing features of different layers.

To prove that high-order interactive concepts are more sensitive to input perturbations than low-order interactive concepts, let us first derive the analytical form of the interaction effect {\small $I(S)$} of an interactive concept.

\begin{lemma}
\label{lemma:taylor_interaction} 
Given a neural network $v$ and an arbitrary input sample {\small $\boldsymbol{x}'\in \mathbb{R}^n$}, 
the network output can be decomposed using the Taylor expansion {\small $v(\boldsymbol{x}') = \sum\nolimits_{S \subseteq N} \sum\nolimits_{\boldsymbol{\pi} \in {Q_S}} U_{S, \boldsymbol{\pi}} \cdot J(S,\boldsymbol{\pi}|\boldsymbol{x}')$}.
In this way, according to Eq.~(\ref{eqn:harsanyi}), the interaction effect {\small $I(S|\bm{x}')$} on the sample $\bm{x}'$ can be reformulated as
\begin{equation}\label{eqn:rewriteI(S)}
\begin{small}
\begin{aligned}
I(S|\boldsymbol{x}')   =
    \sum\nolimits_{\boldsymbol{\pi} \in Q_S} U_{S, \boldsymbol{\pi}} \cdot J(S,\boldsymbol{\pi}|\boldsymbol{x}'),
\end{aligned}
\end{small}
\end{equation}
where  {\small $J(S,\boldsymbol{\pi}|\boldsymbol{x}')=\prod\nolimits_{i \in S} \left({\rm sign}(x'_i - r_i) \cdot \frac{x'_i-{r}_{i}}{\tau}\right)^{\pi_{i}}$} denotes an expansion term of the degree $\boldsymbol{\pi}$, {\small $\boldsymbol{\pi} \in Q_S = \{[\pi_1, \dots, \pi_n]| \forall i \in S, \pi_i \in \mathbb{N}^+; \forall i \not\in S, \pi_i =0\}$}.
{\small $U_{S,\boldsymbol{\pi}} {=}$ $ \frac{\tau^m}{\prod_{i=1}^n \pi_i!} \frac{\partial^{m} v(\boldsymbol{x}_{\emptyset})}{\partial x_{1}^{\pi_{1}} \cdots \partial x_{n}^{\pi_{n}}} \cdot \prod_{i \in S}  [{\rm sign}(x'_i- r_i)]^{\pi_i}$}, {\small $m = \sum_{i=1}^n \pi_i$}.
\end{lemma}

Lemma \ref{lemma:taylor_interaction} provides a new perspective to analyze the sensitivity of the interaction effect {\small $I(S)$}.
In particular, just like in \citet{ren2021AOG} and \citet{ren2023can}, we mask the input variable $x_i$ by setting it to its reference value $x_i \leftarrow r_i$. The reference value $r_i$ is designed as follows.
Let {\small $\mathbb{E}_{\boldsymbol{x}}[x_i]$} denote the average value of the input variable $x_i$ over all input samples, which is usually regarded as a no-information state of this input variable~\cite{ancona2019explaining}. 
In this paper, we remove the information from the input variable $x_i$ by pushing $x_i$ by a large enough distance $\tau$ towards its mean value. In other words, if {\small $x_i > \mathbb{E}_{\boldsymbol{x}}[x_i]$},  
we set the reference value {\small $r_i = x_i - \tau$}\footnote{We need to avoid the case of over-perturbation, by setting { $r_i\leftarrow \max(r_i,\mathbb{E}_{\bm{x}}[x_i])$, if $x_i>\mathbb{E}_{\bm{x}}[x_i]$}; { $r_i\leftarrow \min(r_i,\mathbb{E}_{\bm{x}}[x_i])$}, otherwise. However, such cases are not common in real applications, so we ignore such settings in the following analysis.}; otherwise, {\small $r_i = x_i + \tau$}.
Here, {\small $\tau \in \mathbb{R}$} is a pre-defined constant. 
In this way, compared to setting $r_i = \mathbb{E}_{\boldsymbol{x}}[x_i]$, the above setting ensures comparable perturbation magnitudes over different input dimensions.

Furthermore, in order to simplify the proof, when we add a small Gaussian perturbation {\small $\bm{\epsilon} \sim \mathcal{N}(\bm{0}, \sigma^2\bm{I})$} to the sample {\small $\bm{x}$}, we ignore the extremely low possibility of large perturbations {\small $|\epsilon_i|\ge \tau$} because the variance $\sigma^2$ is small.

Let us start with a simple case in Lemma \ref{lemma:taylor_interaction}. Since people usually adopt low-order Taylor expansion for approximation in real implementations, we first  approximate the interaction effect {\small $I(S|\bm{x}')$} using the expansion term of the lowest degree, and analyze the influence of input perturbations on {\small $I(S|\bm{x}')$}.

\begin{theorem} 
\label{theorem:variance_degree1} 
Let {\small $\boldsymbol{\hat \pi}$} denote the lowest degree of the expansion terms of the interaction effect {\small $I(S|\bm{x}')$}, \textit{i.e.}, {\small $\forall i \in S, \hat \pi_i = 1; \forall i \not\in S, \hat \pi_i= 0$}. 
Let us consider the interaction effect {\small $I(S|\bm{x}')$} only containing the expansion term of the lowest degree, \textit{i.e.}, {\small $I(S|\bm{x}') = U_{S, \bm{\hat \pi}}\cdot J(S, \bm{\hat \pi}|\bm{x}')$}.
In this way, the mean and variance of the interaction effect 
{\small $I(S|\bm{x}'=\boldsymbol{x}+\boldsymbol{\epsilon})$} over different perturbations {\small $\boldsymbol{\epsilon}$} are given as
\begin{equation}
\begin{small}
\begin{aligned}
  \mathbb{E}_{\bm{\epsilon}}[I(S|\bm{x}+\bm{\epsilon})]  &= U_{S,\bm{\hat \pi}}, \\ 
{\rm Var}_{\bm{\epsilon}}[I(S|\bm{x}+\bm{\epsilon})] &= U_{S,\bm{\hat \pi}}^2 ((1+(\sigma/\tau )^2)^{|S|} - 1).
\end{aligned}
\end{small}
\end{equation}
\end{theorem}

Theorem \ref{theorem:variance_degree1} proves that \textit{the variance {\small ${\rm Var}_{\bm{\epsilon}}[I(S|\bm{x}+\bm{\epsilon})]$} increases along with the order {\small $|S|$} of the interactive concept in an exponential manner}. 
It indicates that high-order interactive concepts are much more sensitive to input perturbations than low-order concepts.
Furthermore, as mentioned in Section \ref{sec:approximation}, since we can add perturbations to a surrogate DNN model to well mimic feature representations of a BNN, 
\textbf{we can consider that high-order interactive concepts encoded by the BNN are much more sensitive to weight uncertainty in the BNN than low-order concepts.}

\begin{figure*}\centering
\includegraphics[ width=0.9\textwidth]{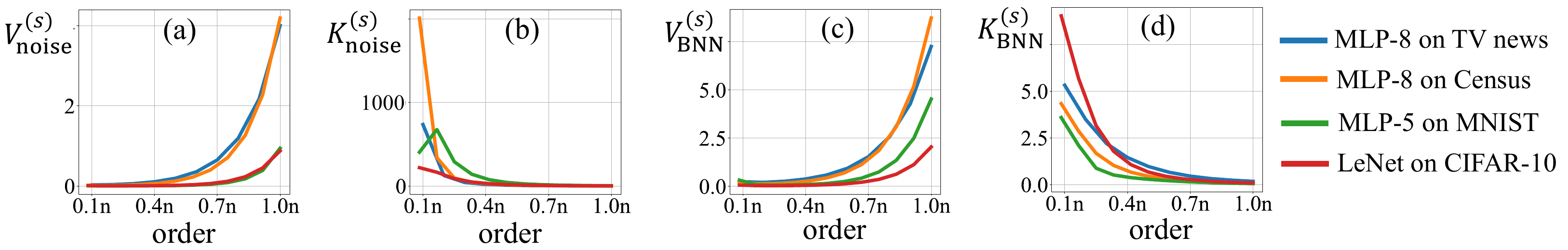}
% \vspace{-0.1cm}
\caption{(a) The exponential increase of the average variance {\small $V^{(s)}_{\text{noise}}$} and (b) the roughly exponential decrease of the average relative stability {\small $K^{(s)}_{\text{noise}}$} along with the order $s$, under perturbations from a distribution {\small $\bm{\epsilon}\sim \mathcal{N}(\bm{0}, 0.05^2\cdot \bm{I})$}. (c) The exponential increase of the average variance {\small $V^{(s)}_{\text{BNN}}$} and (d) the roughly exponential decrease of the average relative stability {\small $K^{(s)}_{\text{BNN}}$} along with the order $s$, under weight uncertainty in the BNN.
}
\label{fig:step2-3-exp}
\end{figure*}

\begin{theorem}[Proof in Appendix \ref{apdx:proof_variance_degree_k}]\label{theorem:variance_degree_k} 
Let {\small $\bm{\pi} \in Q_S = \{[\pi_1, \dots, \pi_n] | \forall i \in S, \pi_i \in \mathbb{N}^+; \forall i \not\in S, \pi_i =0\}$} denote an arbitrary degree. 
Then, the mean and the variance of {\small $J(S, \bm{\pi}|\boldsymbol{{x}}+\boldsymbol{\epsilon})$} over perturbations {\small $\bm{\epsilon}$} are
\begin{equation}
\begin{small}
\begin{aligned}
  \mathbb{E}_{\bm{\epsilon}}[J(S,\bm{\pi}|\bm{{x}}+\bm{\epsilon})] &= \mathbb{E}_{\bm{\epsilon}}[\prod\nolimits_{i \in S}(1 + \frac{\epsilon_i}{\tau})^{\pi_i}], \\
{\rm Var}_{\bm{\epsilon}}[J(S,\bm{\pi}|\bm{{x}}+\bm{\epsilon})] &= {\rm Var}_{\bm{\epsilon}}[\prod\nolimits_{i \in S}(1 + \frac{\epsilon_i}{\tau})^{\pi_i}]
\end{aligned}
\end{small}
\end{equation}
\end{theorem}

Theorem \ref{theorem:variance_degree_k} extends Theorem \ref{theorem:variance_degree1} to a \textbf{general case}, where we use a higher-order Taylor expansion to represent {\small $I(S|\bm{x}')$}.

\begin{theorem}[Proof in Appendix \ref{apdx:proof_high_large_variance}]\label{theorem:high_large_variance} 
Let {\small $S$ and $S'$} be two interactive concepts, such that {\small $S \subsetneq S'$}. 
Let us consider expansion terms {\small $J(S,\bm{\pi})$} and {\small $J(S',\bm{\pi}')$},
where the term {\small $J(S',\bm{\pi}')$} is extended from the term {\small $J(S,\bm{\pi})$} with {\small $\bm{\pi} \prec \bm{\pi}^\prime$}. I.e., (1) {\small $\forall i \in S', \pi'_i \in \mathbb{N}^+$}; otherwise, {\small $\pi'_i = 0$}. (2) Given {\small $\bm{\pi}'$}, {\small $\forall j \in S, \pi_j = \pi'_j$}; otherwise, {\small $\pi_j = 0$}.
Then, we have 
\begin{equation}
\begin{small}
\begin{aligned}
% --- first inequality starts
& \frac{{\rm Var}_{\bm{\epsilon}}[J(S', \bm{\pi}'|\bm{{x}}+\bm{\epsilon})]}
{{\rm Var}_{\bm{\epsilon}}[J(S,\bm{\pi}|\bm{{x}}+\bm{\epsilon})]}  > 
\prod\nolimits_{i \in S' \setminus S}  \mathbb{E}^2_{\epsilon_i}[(1 + \frac{\epsilon_i}{\tau})^{\pi'_i}], \\
& \frac
{\mathbb{E}_{\bm{\epsilon}}[J(S',\bm{\pi}'|\bm{{x}}+\bm{\epsilon})]/
{\rm Var}_{\bm{\epsilon}}[J(S', \bm{\pi}'|\bm{{x}}+\bm{\epsilon})]}
{\mathbb{E}_{\bm{\epsilon}}[J(S,\bm{\pi}|\bm{{x}}+\bm{\epsilon})]/{\rm Var}_{\bm{\epsilon}}[J(S,\bm{\pi}|\bm{{x}}+\bm{\epsilon})]}\\
&\quad \quad \quad \quad \quad \quad \quad \quad  \quad  <
\frac{1}{\prod\nolimits_{i \in S' \setminus S}  \mathbb{E}_{\epsilon_i}[(1 + \frac{\epsilon_i}{\tau})^{\pi'_i}]},
\end{aligned}
\end{small}
\end{equation}
and we can also obtain {\small $\mathbb{E}_{\epsilon_i}[(1 + \frac{\epsilon_i}{\tau})^{\pi'_i}] \ge 1$}.
\end{theorem}

Theorem \ref{theorem:high_large_variance} indicates that for an arbitrary degree {\small $\bm{\pi}$} of the interactive concept {\small $\bm{S}$}, 
{\small ${\rm Var}_{\bm{\epsilon}}[J(S', \bm{\pi}'|\bm{{x}}+\bm{\epsilon})]/{\rm Var}_{\bm{\epsilon}}[J(S,\bm{\pi}|\bm{{x}}+\bm{\epsilon})]$} increases in an exponential manner along with  {\small $|S' \setminus S|=|S'|-|S|$}. 
Therefore, we can roughly consider that {\small ${\rm Var}_{\bm{\epsilon}}[J(S,\bm{\pi}|\bm{{x}}+\bm{\epsilon})]$} increases exponentially \textit{w.r.t.} the order {\small $|S|$}. 
Furthermore, according to Lemma \ref{lemma:taylor_interaction}, {\small $I(S|\bm{x}+\bm{\epsilon})$} can be re-written as the weighted sum of {\small $J(S,\bm{\pi}|\bm{x}+\bm{\epsilon})$}. Since coefficients {\small $U_{S,\bm{\pi}}$} \textit{w.r.t.} different {\small $S$} and {\small $\bm{\pi}$} are usually chaotic, 
we can roughly consider that the sensitivity of {\small $I(S|\bm{x} + \bm{\epsilon})$}  also grows exponentially along with the order {\small $|S|$} of the interactive concept {\small $S$}. 
In addition, Theorem \ref{theorem:high_large_variance} also proves the approximately exponential decrease of {$\frac{\mathbb{E}_{\bm{\epsilon}}[J(S',\bm{\pi}'|\bm{{x}}+\bm{\epsilon})]
/{\rm Var}_{\bm{\epsilon}}[J(S',\bm{\pi}'|\bm{{x}}+\bm{\epsilon})]}
{\mathbb{E}_{\bm{\epsilon}}[J(S,\bm{\pi}|\bm{{x}}+\bm{\epsilon})]/{\rm Var}_{\bm{\epsilon}}[J(S, \bm{\pi}|\bm{{x}}+\bm{\epsilon})]}$} along with {\small  $|S'|-|S|$}.
Similarly, we can obtain that the relative stability {\small $\mathbb{E}_{\bm{\epsilon}}[I(S|\bm{{x}}+\bm{\epsilon})]
/{\rm Var}_{\bm{\epsilon}}[I(S|\bm{{x}}+\bm{\epsilon})]$} decreases along with 
the order {\small $|S|$}.

\textbf{Conclusions.} Both Theorem \ref{theorem:variance_degree1} and Theorem \ref{theorem:high_large_variance} tell us that high-order interactive concepts are much more sensitive to input perturbations. Furthermore, combined with the conclusion in Section \ref{sec:approximation}, 
\textbf{we can conclude that high-order interactive concepts encoded by the BNN are much more sensitive to the weight uncertainty in the BNN than low-order concepts.}

\textbf{Experimental verification.} 
We conducted experiments to verify the above conclusions.
To verify the sensitivity to input perturbations, we added a random perturbation {\small $\bm{\epsilon}\sim \mathcal{N}(\bm{0},\sigma^2\bm{I})$} to a given input sample {\small $\bm{x}$}, where {\small $\sigma^2=0.05^2$}. 
Then, we used the following two metrics,  
{\small $V_{\text{noise}}^{(s)} =$} {\small $\mathbb{E}_{\bm{x}}[\mathbb{E}_{|S|=s}[{\rm Var}_{\bm{\epsilon} \sim \mathcal{N}(\bm{0},\sigma^2\bm{I})}[I(S|\bm{x}+\bm{\epsilon})]]]$} and 
{\small $K_{\text{noise}}^{(s)}=$ $\mathbb{E}_{\bm{x}}[\mathbb{E}_{|S|=s}
[\frac{|\mathbb{E}_{\bm{\epsilon} \sim \mathcal{N}(\bm{0},\sigma^2\bm{I})}[I(S|\bm{x}+\bm{\epsilon}) ]|}    {{\rm Var}_{\bm{\epsilon} \sim \mathcal{N}(\bm{0},\sigma^2\bm{I})}[I(S|\bm{x}+\bm{\epsilon})]}]]$}, to measure the average variance and the average relative stability of the $s$-order interactive concepts \textit{w.r.t.} the input perturbation {\small $\bm{\epsilon}$}. 
Then, a large  {\small $V_{\text{noise}}^{(s)}$} or a small  {\small $K_{\text{noise}}^{(s)}$} indicated that the $s$-order interactive concepts were sensitive to input perturbations.

Similarly, to verify the sensitivity to the weight uncertainty, we sampled different weights {\small $\bm{W}$} from the weight distribution {\small $q_{\bm{\theta}}(\bm{W})$} of the BNN. Then, we used   
{\small $V_{\text{BNN}}^{(s)} = \mathbb{E}_{\bm{x}}[\mathbb{E}_{|S|=s}[{\rm Var}_{\bm{W} \sim q_{\bm{\theta}}(\bm{W})}[I(S|\bm{x}, \bm{W})]]]$} and 
{\small $K_{\text{BNN}}^{(s)} = $ $\mathbb{E}_{\bm{x}}[\mathbb{E}_{|S|=s}
[\frac{|\mathbb{E}_{\bm{W} \sim q_{\bm{\theta}}(\bm{W})}[I(S|\bm{x}, \bm{W}) ]|}   {{\rm Var}_{\bm{W} \sim q_{\bm{\theta}}(\bm{W})}[I(S|\bm{x}, \bm{W})]}]]$} to measure the average variance and the average relative stability of the $s$-order interactive concepts \textit{w.r.t.} the weight uncertainty in the BNN. 
Therefore, a large value of {\small $V_{\text{BNN}}^{(s)}$} or a small value of {\small $K_{\text{BNN}}^{(s)}$} indicated that the $s$-order interactive concepts were sensitive to the weight uncertainty. We followed experimental settings in the \textit{experiments} paragraph in Section \ref{sec:approximation} to train BNNs.
Specifically, we trained BNNs with the MLP architecture  on the MNIST dataset, the TV news dataset, and the Census dataset. We trained BNNs with the LeNet architecture on the CIFAR-10 dataset. 
Appendix \ref{sec:apdx_exp_detail} introduces how to efficiently compute $I(S|\bm{x})$ on images.

Figure \ref{fig:step2-3-exp} shows that the average variance {\small $V_{\text{noise}}^{(s)}$} and {{\small $V_{\text{BNN}}^{(s)}$}} increased exponentially along with the order $s$, while the relative stability {\small $K_{\text{noise}}^{(s)}$} and {{\small $K_{\text{BNN}}^{(s)}$}} both decreased along with the order. 
This demonstrated that high-order interactive concepts were much more sensitive to input perturbations and the weight uncertainty in the BNN, thereby verifying Theorem \ref{theorem:variance_degree1} and Theorem \ref{theorem:high_large_variance}.

\subsection{Perturbation-sensitive concepts are difficult to learn}\label{sec:learningdifficulty}
In this subsection, we prove that high-order interactive concepts, which are sensitive to input perturbations and weight uncertainty, are difficult to be learned by a BNN in a regression task. Specifically, we measure the learning effects of interactive concepts (denoted by {\small$U_S$}), and Theorems \ref{theorem:w_relationship} and \ref{theorem:C_Supperbound} prove the small learning effects of perturbation-sensitive concepts.

To facilitate the analysis, we first simplify the conceptual learning as a linear problem.
Specifically, we first rewrite the interaction effect of an interactive concept {\small $S$}. 
Given an input sample $\bm{x}$, according to Eq.~(\ref{eqn:rewriteI(S)}), the interaction effect of the concept {\small $S$} on the sample $\bm{x}'$ (obtained by applying some transformations on $\bm{x}$),
{\small $I(S|\bm{x}')$},  can be rewritten as
\begin{equation}
 \begin{small}
  \begin{aligned}
  \label{eqn:def_US}
I(S|\bm{x}') = U_S \cdot C_S(\bm{x}'),  
  \end{aligned}
 \end{small}
\end{equation}
where  the constant {\small $U_S = I(S|\bm{x})$} denotes the interaction effect of the concept {\small $S$},
and the function for the activation state is given as {\small $C_S(\bm{x}') =\sum\nolimits_{\boldsymbol{\pi} \in Q_S}$ $U_{S, \boldsymbol{\pi}} J(S,\boldsymbol{\pi}|\boldsymbol{x}')/U_S$}.

\textbf{Understanding of $C_S(\bm{x}')$.}
Let us consider a sample {\small $\bm{x}'$} where each input variable $x'_i$ is either masked by the reference value $r_i$ or kept unchanged as $x_i$. Then, the function {\small $C_S(\bm{x}')$} defined above represents the binary activation state of the concept {\small $S$} in the sample {\small $\bm{x}'$}, which is an AND relationship between all variables in {\small $S$}:
\begin{equation}
\label{eqn:C_Sequivalent}
\begin{small}
\begin{aligned}
C_S(\bm{x}')  = \bigwedge\nolimits_{i\in S} {\rm unmask}({x}'_i),
\end{aligned}
\end{small}
\end{equation}
where the binary function {\small ${\rm unmask}({x}'_i) \in \{0,1\}$} checks whether the $i$-th variable {\small${x}'_i$} is masked in the sample {\small $\bm{x}'$}. If the $i$-th variable is masked, then {\small${\rm unmask}({x}'_i)=0$}; otherwise, {\small${\rm unmask}({x}'_i)=1$}.

Only when all input variables in $S$ are not masked in the sample {\small$\bm{x}'$}, the concept {\small$S$} is activated, and  {\small$C_{S}(\bm{x}') =1$}. 
If any input variable in {\small $S$} is masked, then the concept {\small$S$} will not be activated ({\small$C_{S}(\bm{x}') =0$}), yielding zero interaction effect {\small $I(S|\bm{x}') = 0$}.

Thus, we can extend Eq.~(\ref{eqn:effciency}) to a continuous version that explains the output as a \textbf{linear regression problem}. 
\begin{equation}
 \begin{small}
  \begin{aligned}
  \label{eqn:efficiencynew}
 v(\bm{x}') & = \sum\nolimits_{S \in \Omega} U_S \cdot C_S(\bm{x}'), \\
           \end{aligned}
 \end{small}
\end{equation}
where the activation state {\small$C_S(\bm{x}')$} can be considered as an input dimension of the linear function, which reflects whether the input sample {\small$\bm{x}'$}  contains the  concept {\small$S$}.

Therefore, the absolute value of the coefficient {\small $U_S$} can be considered as \textit{the strength of the neural network in learning the interactive concept {\small $S$}}.
According to Section \ref{sec:representing} and~\citet{ren2021AOG}, most interactive concepts have negligible coefficients {\small $|U_S| \approx 0$},
so we can consider that the neural network only encodes a few interactive concepts {\small$S$} with large absolute values {\small$|U_S|$}.

\begin{figure*}[t]
\centering
\includegraphics[width=0.9\textwidth]{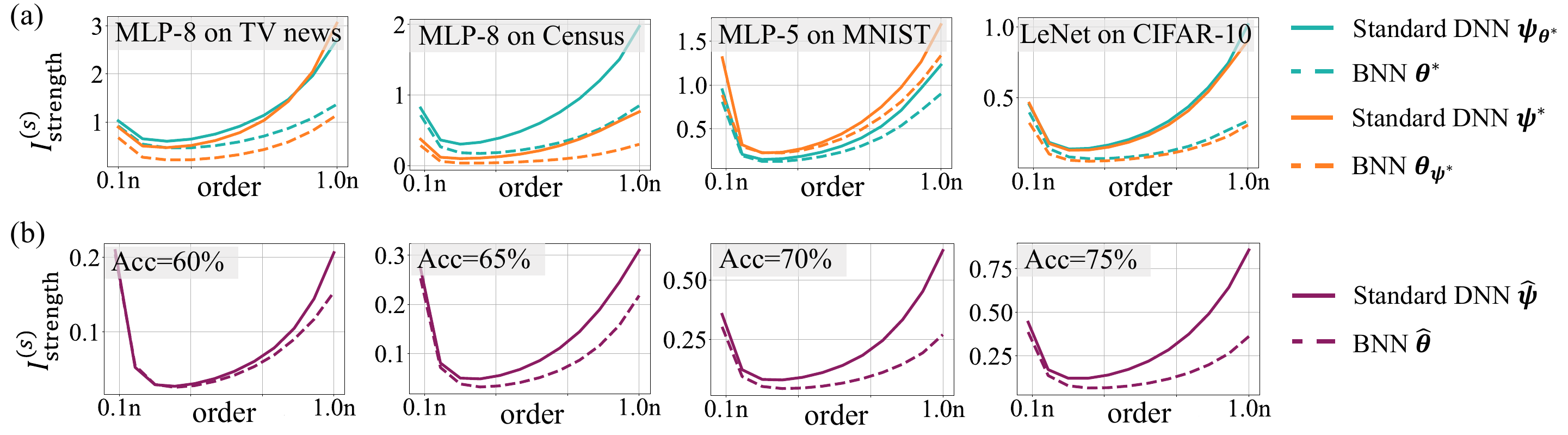}
% \vspace{-0.1cm}
\caption{(a) Comparison of the strength of interactive concepts  (i) between a trained BNN {\small $\bm{\theta}^*$} and the constructed standard DNN {\small $\bm{\psi}_{\bm{\theta}^*}$}, (ii) between a trained standard DNN {\small $\bm{\psi}^*$} and the constructed BNN {\small $\bm{\theta}_{\bm{\psi}^*}$}. (b) We trained a standard DNN {\small $\hat{\bm{\psi}}$} and a BNN {\small $\hat{\bm{\theta}}$} with the LeNet architecture on the CIFAR-10 dataset, and compared the strength of interactive concepts between the two networks when the two networks were trained to have the same training accuracy.}
\label{fig:BNNvsDNN}
% \vspace{-0.1cm}
\end{figure*}
%%%%%%%%%%%%%%%%%%%%%%%%%%%%%%%%%%%%%%

Let us facilitate the poof on a regression task. Based on the conclusion in Section \ref{sec:approximation}, we can roughly consider that training a BNN on normal samples is equivalent to training a surrogate DNN model on perturbed input samples  {\small $\bm{x}'=\bm{x}+\bm{\epsilon}$}. 
Then, according to Eq.~(\ref{eqn:efficiencynew}), the learning of the BNN on a certain input sample can be roughly represented as {\small $\min_{\{U_S|S\in \Omega\}} L(\{U_S\})$}, and the loss is given by
\begin{equation}\label{eqn:MSEloss2}
\begin{small}
\begin{aligned}
  L(\{U_S\})
&= \mathbb{E}_{\bm{\epsilon}} \left[(y^* - v(\bm{x}'))^2 \right]\\
&= \mathbb{E}_{\bm{\epsilon}} [(y^* - \sum\nolimits_{S\in \Omega} U_S \cdot C_S(\bm{x}+\bm{\epsilon}))^2 ]
\end{aligned}
\end{small}
\end{equation}
where $\bm{x}$ and $y^*$ denote the input sample and the ground-truth output, respectively, and {\small $\bm{x}'=\bm{x}+\bm{\epsilon}$}.

\begin{theorem}[Proof in Appendix \ref{apdx:proof_w_relationship}]\label{theorem:w_relationship} 
Given two random interactive concepts {\small $S$} and {\small $S'$}, we can roughly assume that  {\small $C_S(\bm{x}+\bm{\epsilon})$} is independent of {\small $C_{S'}(\bm{x}+\bm{\epsilon})$}, because the two concepts {\small$S$} and {\small$S'$} usually have little overlap in most cases. Let {\small$\mathbb{E}_{\bm{\epsilon}} [C_S(\bm{x}+\bm{\epsilon})]$} and {\small${\rm Var}_{\bm{\epsilon}} [C_S(\bm{x}+\bm{\epsilon})]$} denote the mean and the variance of  {\small $C_S(\bm{x}+\bm{\epsilon})$} w.r.t. {\small $\bm{\epsilon}$}, respectively. 
Then, the solution  to Eq.~(\ref{eqn:MSEloss2}) satisfies the following property:
\begin{equation}
\begin{small}
\begin{aligned}
      \forall \ S \in \Omega, \quad {|U^*_S|} \propto \vert \mathbb{E}_{\bm{\epsilon}} [C_S(\bm{x}+\bm{\epsilon})] / {\rm Var}_{\bm{\epsilon}} [C_S(\bm{x}+\bm{\epsilon})]\vert
\end{aligned}
\end{small}
\end{equation}
\end{theorem}
Theorem \ref{theorem:w_relationship} proves that the learning effect of an interactive concept {\small $S$}, measured by {\small $|U^*_S|$}, is proportional to the relative stability of the activation state of the interactive concept {\small $\vert \mathbb{E}_{\bm{\epsilon}} [C_S(\bm{x}+\bm{\epsilon})] / {\rm Var}_{\bm{\epsilon}} [C_S(\bm{x}+\bm{\epsilon})] \vert$} \textit{w.r.t.} perturbations {\small $\bm{\epsilon}$}.
This indicates that perturbation-sensitive interactive concepts are more difficult to learn.

\begin{theorem}[Proof in Appendix \ref{apdx:proof_C_Supperbound}]\label{theorem:C_Supperbound} 
Let {\small $A^{\text{min}} = \min_S |U_S|$}
and {\small $A^{\text{max}} = \max_S |U_S|$} denote the lower bound and the upper bound of {\small $\vert U_S \vert$} over all interactive concepts {\small $S$}. 
Then,  for any {\small $S \subseteq N$}, we have
\begin{equation}\label{eqn:C_Supperbound}
\begin{small}
\begin{aligned}
  A^{\text{min}} \cdot  \frac{ \vert \mathbb{E}_{\bm{\epsilon}} [I(S|\bm{x}+\bm{\epsilon})]\vert} {{\rm Var}_{\bm{\epsilon}} [I(S |\bm{x}+\bm{\epsilon})]} 
  &\leq 
  \frac{\vert \mathbb{E}_{\bm{\epsilon}} [C_S(\bm{x}+\bm{\epsilon})]\vert} {{\rm Var}_{\bm{\epsilon}} [C_S(\bm{x}+\bm{\epsilon})]} \\
  &\leq  
  A^{\text{max}} \cdot  \frac{ \vert \mathbb{E}_{\bm{\epsilon}} [I(S|\bm{x}+\bm{\epsilon})]\vert} {{\rm Var}_{\bm{\epsilon}} [I(S |\bm{x}+\bm{\epsilon})]}
\end{aligned}
\end{small}
\end{equation}
\end{theorem}

Theorem \ref{theorem:C_Supperbound} proves that high-order (complex) interactive concepts have low relative stability \textit{w.r.t.} perturbations {\small $\bm{\epsilon}$}. 
In fact, both Theorem \ref{theorem:high_large_variance} and Figure \ref{fig:step2-3-exp} have told us that {\small $\vert \mathbb{E}_{\bm{\epsilon}} [I(S|\bm{x}+\bm{\epsilon})] / {\rm Var}_{\bm{\epsilon}} [I(S |\bm{x}+\bm{\epsilon})] \vert$} significantly decreases along with the order {\small $s = \vert S \vert$} of the interactive concept {\small $S$}. 
Therefore, both  the lower bound and the upper bound of {\small $\vert \mathbb{E}_{\bm{\epsilon}} [C_S(\bm{x}+\bm{\epsilon})] / {\rm Var}_{\bm{\epsilon}} [C_S(\bm{x}+\bm{\epsilon})] \vert$} in Eq.~(\ref{eqn:C_Supperbound}) decrease along with the order {\small $s$} significantly.
In this way, we can approximately consider that the strength of encoding a concept {\small $\vert U^*_S \vert \propto \vert \mathbb{E}_{\bm{\epsilon}} [C_S(\bm{x}+\bm{\epsilon})] / {\rm Var}_{\bm{\epsilon}} [C_S(\bm{x}+\bm{\epsilon})]\vert$} also decreases along with the order of interactive concepts. In other words, we prove that high-order interactive concepts are more difficult to be learned under perturbations {\small $\bm{\epsilon}$}. Combining the conclusion in Section \ref{sec:approximation}, we also prove that high-order interactive concepts are more difficult to be learned by the BNN.

%=================================================

\section{Experiments}\label{sec:experiments}
In this section, we experimentally verified that compared to standard DNNs, BNNs were less likely to encode high-order (complex) interactive concepts. Specifically, we constructed three pairs of baseline networks for comparison.

(1) Given a trained BNN {\small $\bm{\theta}^*$}, we constructed a standard DNN  by setting its weights to the mean value of the weight distribution of the BNN. 
The standard DNN was denoted by {\small $\bm{\psi}_{\bm{\theta}^*}$}. Then, we compared the strength of all high-order interactive concepts between the BNN {\small$\bm{\theta}^*$} and the standard DNN {\small $\bm{\psi}_{\bm{\theta}^*}$} without weight/feature uncertainty.

(2) Similarly, given a trained standard DNN {\small $\bm{\psi}^*$}, we constructed a BNN {\small $\bm{\theta}_{\bm{\psi}^*}$} by setting the mean value of its weight distribution to the weights of the standard DNN. 
We set  all weight dimensions in the $l$-th layer of the BNN to share the same variance ${\sigma}_l^2$, where ${\sigma}_l^2$ was computed as the average of variances of all weight dimensions in the $l$-th layer of the previous BNN {\small $\bm{\theta}^*$}. 
Then, we compared the strength of high-order interactive concepts between the standard DNN {\small $\bm{\psi}^*$} and the BNN {\small $\bm{\theta}_{\bm{\psi}^*}$}.

(3) We trained a standard DNN and a BNN with the same architecture. Then, we compared the strength of high-order interactive concepts between each pair of standard DNN   {\small $\hat{\bm{\psi}}$} and the BNN  {\small $\hat{\bm{\theta}}$} when these two networks were trained to have the same training accuracy. We used the training accuracy to align the learning progress of the two networks for fair comparison.

Specifically, the average strength of the $s$-order interactive concepts was measured as {\small $I^{(s)}_{\text{strength}} = \mathbb{E}_{\bm{x}}[\mathbb{E}_{S\subseteq N, |S|=s} [\vert I(S|\bm{x}) \vert]]$}. 
To compute the interaction effect {\small $I(S|\bm{x})$}, we set {\small $v(\bm{x}_S)=\log \frac{p(y=y^*|\bm{x}_S)}{1- p(y=y^*|\bm{x}_S)}\in \mathbb{R}$}, which reflected the confidence of classifying the masked input sample {\small $\bm{x}_S$} into the ground-truth category $y^*$.
For standard DNNs,  {\small $p(y=y^*|\bm{x}_S)$} referred to the classification probability of the ground-truth category on the masked sample {\small $\bm{x}_S$}. For BNNs, {\small $p(y=y^*|\bm{x}_S)$} was computed according to Eq.~(\ref{eq:BNN-inference}), where we 
sampled ten neural networks from the weight distribution {\small $q_{\bm{\theta}}(\bm{W})$} of the BNN, and computed the average classification probability over all these networks.

We followed experimental settings in the  \textit{experiments} paragraph in Section \ref{sec:approximation} to train the networks. Specifically, we trained standard DNNs and BNNs with the MLP architecture on the TV news dataset, the Census dataset, and the MNIST dataset. We trained standard DNNs and BNNs  with the LeNet architecture on the CIFAR-10 dataset. Appendix \ref{sec:apdx_exp_detail} introduces how to efficiently compute $I(S|\bm{x})$ on images.
Figure \ref{fig:BNNvsDNN} shows that the strength of high-order interactive concepts of BNNs was much weaker than that of standard DNNs in all comparisons. This verified that BNNs were less likely to encode high-order (complex) interactive concepts than standard DNNs.

%======================================

\section{Conclusion and discussion}
In this paper, we have proven the tendency of mean-field variational BNNs to avoid encoding high-order (complex)  concepts. Many studies~\cite{ren2021AOG,li2023does, ren2023where} have shown that there does exist a concept-emerging phenomenon when a neural network is sufficiently trained. 

Besides, as discussed in the introduction, encoding less complex concepts does not mean that BNNs have weaker representation power than standard DNNs, because a standard DNN can be considered as a specific BNN with zero weight uncertainty. More crucially,~\citet{ren2021towards} and~\citet{pmlr-v151-lengerich22a} proved that high-order concepts  are usually vulnerable to adversarial attacks and have weak generalization power. Thus, it is hard to say whether the tendency to avoid encoding complex concepts is a demerit or not.

\paragraph{Acknowledgements.} This work is partially supported by the National Nature Science Foundation of China (62276165), National Key R\&D Program of China (2021ZD0111602), Shanghai Natural Science Foundation (21JC1403800,21ZR1434600), National Nature Science Foundation of China (U19B2043). This work is also partially supported by Huawei Technologies Inc.

% \newpage

\bibliography{example_paper}
\bibliographystyle{icml2023}

%%%%%%%%%%%%%%%%%%%%%%%%%%%%%%%%%%%%%%%%%%%%%%%%%%%%%%%%%%%%%%%%%%%%%%%%%%%%%%%
%%%%%%%%%%%%%%%%%%%%%%%%%%%%%%%%%%%%%%%%%%%%%%%%%%%%%%%%%%%%%%%%%%%%%%%%%%%%%%%
% APPENDIX
%%%%%%%%%%%%%%%%%%%%%%%%%%%%%%%%%%%%%%%%%%%%%%%%%%%%%%%%%%%%%%%%%%%%%%%%%%%%%%%
%%%%%%%%%%%%%%%%%%%%%%%%%%%%%%%%%%%%%%%%%%%%%%%%%%%%%%%%%%%%%%%%%%%%%%%%%%%%%%%
\newpage
\appendix
\onecolumn

\section{Discussion on literature in representation capacities of BNNs}
Many studies investigated the representation capacity of BNNs from different perspectives.~\citet{gal2018sufficient} and~\citet{carbone2020robustness} proved that BNNs were robust to adversarial attacks.~\citet{kristiadi2020being} proved that BNNs could mitigate the over-confidence problem in standard ReLU networks.~\citet{wenzel2020good} considered that the poor performance of BNNs was due to the inappropriate prior distribution of weights in the BNN, and a series of studies~\cite{wu2018deterministic, krishnan2020specifying, fortuin2022bayesian} found that using carefully-designed prior distributions of weights could improve the performance of the BNN.~\citet{zhang2022improving} also showed that adding adversarial perturbations to weights during training could improve the performance of the BNN.
Besides,~\citet{foong2020expresiveness} proved that using either fully-factorized Gaussian distributions or dropout operations to approximate the posterior distribution of a BNN would lead to inaccurate  uncertainty estimation of the network prediction.
Unlike previous studies, we focus on the conceptual representation of BNNs, and theoretically prove that mean-field variational BNNs are less likely to encode complex interactive concepts than standard DNNs.

\section{Discussion on literature in interactions in neural networks}  
Interactions in game theory are often used to explain neural networks and are closely related to the quantification of concepts.~\citet{grabisch1999axiomatic} first proposed the Shapley interaction index, and~\citet{lundberg2018consistent} later used this index to explain tree ensembles.~\citet{janizek2021explaining} explained the pairwise feature interaction in DNNs, while~\citet{sundararajan2020shapley} proposed the Shapley Taylor interaction index to quantify interactions among multiple input variables.~\citet{ren2021AOG} used game-theoretic interactions to analyze the emergence of concepts in the training of neural networks, and proved the faithfulness and sparsity of such formulation of concepts.
In this paper, we follow the definition of concepts in~\citet{ren2021AOG}, and prove BNNs' tendency to avoid encoding high-order (complex) concepts.

\section{Discussion on literature in the connection between adversarial robustness and interpretability}  
Recent studies have shown that adversarial robustness is closely related to the interpretability of neural networks. \citet{etmann2019on} discovered and explained the phenomenon that adversarially robust models exhibit simpler and more human-interpretable saliency maps. \citet{engstrom2019adversarial} demonstrated that adversarially robust models showed clear human-recognizable features when using the optimization-based feature visualization method \cite{olah2017feature}, and the mapping from input images to intermediate features of the model is approximately invertible. \citet{ilyas2019adversarial} demonstrated that adversarial samples can be attributed to the existence of non-robust features (features that are noisy and not interpretable to humans, but are highly predictive). \citet{ren2021towards}  showed that high-order (complex) interactive concepts encoded by neural networks are vulnerable to adversarial attacks, and that adversarially-trained DNNs encode more discriminative low-order (simple) interactive concepts than standard DNNs. In this paper, we prove that BNNs tend to avoid encoding high-order (complex) interactive concepts, which implies that BNNs may exhibit good adversarial robustness, from the perspective of conceptual representations.

\newpage

\section{Experiments on the connection between conceptual complexity and adversarial robustness}\label{apdx:exp_complexity_vs_robustness}
\begin{figure}[h]
    \centering
    \includegraphics[width=0.9\linewidth]{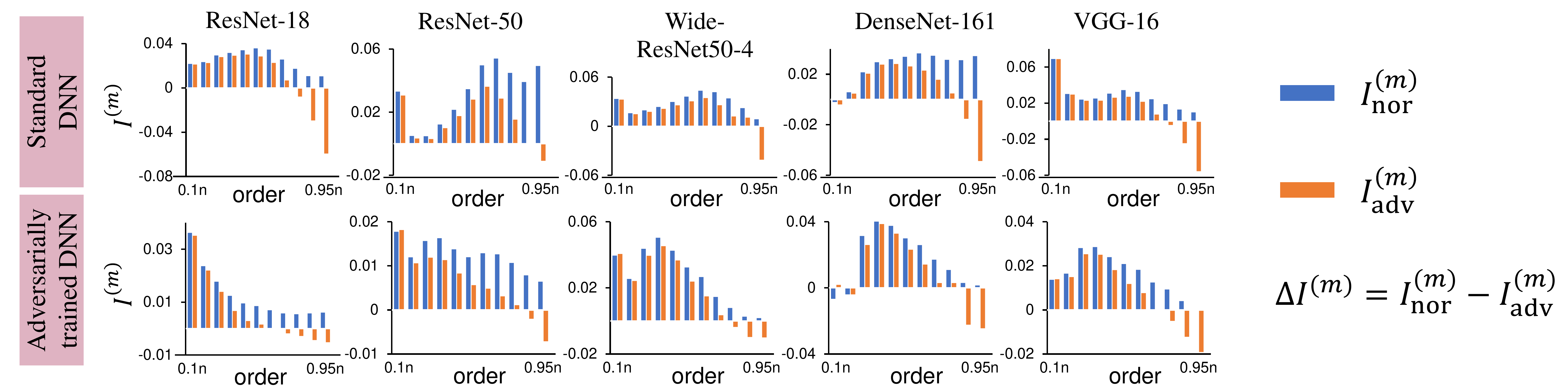}
    \vspace{-0.2cm}
    \caption{Adversarial attacks mainly affect high-order interactive concepts. Please refer to \citet{ren2021towards} for more details.}
    \label{fig:robustness}
\end{figure}

We show experimental results in~\cite{ren2021towards} to demonstrate that high-order (complex) interactive concepts are more vulnerable to adversarial attacks, as illustrated in Figure \ref{fig:robustness}. 
Although the interaction used in ~\cite{ren2021towards} was a bit different from the interaction used in this paper, we can prove that the Harsanyi dividend interaction in this paper is the elementary component of the multi-order interaction in ~\cite{ren2021towards}. Thus, experimental results still reflect adversarial vulnerability of high-order interactive concepts.
Please see \citet{ren2021towards} for more details.

\section{Comparison of adversarial robustness between BNNs and standard DNNs} 
\label{apdx:exp_adv_robustness}
This experiment compares the adversarial robustness between BNNs and standard DNNs. Specifically, we train two BNNs with 8-layer MLP architecture on tabular datasets, including the Census dataset and the TV news dataset. All MLPs contain 100 neurons in each hidden layer. For each trained BNN $\bm{\theta}^*$, we compare this BNN with a standard DNN $\bm{\psi}_{\bm{\theta}^*}$ that is constructed by following the {experimental setting} of \textbf{Comparison (1)} in Section \ref{sec:experiments}. In other words, the standard DNN $\bm{\psi}_{\bm{\theta}^*}$ is constructed by setting its weights to the mean value of the weight distribution of the BNN. Thus, with such experimental settings, the main difference between the BNN $\bm{\theta}^*$ and the DNN $\bm{\psi}_{\bm{\theta}^*}$  is the weight uncertainty of BNNs, so that our experiment can faithfully reflect the impact of weight uncertainty of BNNs on the adversarial robustness.

We compare the classification accuracy on adversarial samples in the testing set between the BNN $\bm{\theta}^*$ and the standard DNN $\bm{\psi}_{\bm{\theta}^*}$. To this end, for each pair of BNN and standard DNN, we adopt the untargeted PGD adversarial attack~\citep{madry2018towards} based on the $l_\infty$ norm, and accordingly obtain their accuracies on adversarial samples. In the PGD attack based on the $l_\infty$ norm, an adversarial sample $\tilde{\bm{x}}$ is constrained wthin the $l_\infty$-ball around the original sample $\bm{x}$, \textit{i.e.}, $\Vert \tilde{\bm{x}} - {\bm{x}} \Vert_\infty \le \epsilon$. We conduct the attack for 20 steps with $\epsilon=0.1$, and set the step size to 0.01. Table \ref{tab:adv_robustness} shows that BNNs exhibit higher adversarial accuracies than the corresponding DNNs, which indicates that BNNs are more robust to adversarial attacks.

\begin{table}[h]
\centering
\caption{Adversarial accuracies of the BNN ${\bm{\theta}^*}$ and the standard DNN $\bm{\psi}_{\bm{\theta}^*}$
 constructed based on the BNN.}
 \vspace{0.2cm}
\begin{tabular}{lcccc} 
\toprule
   & MLP-8 on Census & MLP-8 on TV news \\ 
   \hline
BNN ${\bm{\theta}^*}$ & \textbf{77.51\%} &  \textbf{53.54\%} \\
DNN $\bm{\psi}_{\bm{\theta}^*}$ & 75.22\% & 50.54\% \\ 
\bottomrule
\end{tabular}
\label{tab:adv_robustness}
\end{table}

\section{Experiments on the connection between conceptual complexity and generalization power.}
\label{apdx:exp_complexity_vs_generalization}
~\citet{zhou2023concept} have investigated the connection between the generalization ability of an interactive concept encoded by a neural network and the complexity (order) of this concept.

The {generalization ability of a concept} $S$ is defined as follows. For a generalizable concept $S$, if the concept is frequently extracted from training samples, then it is supposed to be also frequently extracted from testing samples and to make consistently positive (or consistently negative) effects to the classification of a certain category. Otherwise, this concept would not be considered generalizable. Thus, the generalization ability of a specific concept can be evaluated by whether this concept's interaction effects over training samples are similar to its interaction effects over testing samples. 
To this end, ~\citet{zhou2023concept} quantified the \textit{average generalization ability} $g^{(m)}$ over all $m$-order interactive concepts by the similarity between interaction effects of $m$-order interactive concepts in training samples and those in testing samples:
\begin{equation}
    g^{(m)} \overset{\text{\rm def}}{=} \mathbb{E}_{c}\left[\operatorname{sim}\left(I_{\text {train, c}}^{(m)}, I_{\text {test }, c}^{(m)}\right)\right],
\end{equation}
where the vector $I_{\text{train, c}}^{(m)} =\left[\mathbb{E}_{\bm{x} \in \text{train}, c}\left[I\left(S_{1} | \bm{x}\right)\right], \mathbb{E}_{\bm{x} \in \text{train}, c}\left[I\left(S_{2} | \bm{x}\right)\right], \ldots, \mathbb{E}_{\bm{x} \in \text{train}, c}\left[I\left(S_{d} | \bm{x}\right)\right]\right]^{\top} \in \mathbb{R}^d$ denotes interaction effects of all $m$-order interactive concepts $[S_1, \cdots, S_d]$. The interaction effect of each concept 
$\mathbb{E}_{\bm{x} \in \text{train},c}\left[I\left(S_{i} | \bm{x}\right)\right]$ is averaged over different training samples in the category $c$.
Accordingly, the vector $I_{\text{test, c}}^{(m)} =\left[\mathbb{E}_{\bm{x} \in \text{test}, c}\left[I\left(S_{1} | \bm{x}\right)\right], \mathbb{E}_{\bm{x} \in \text{test}, c}\left[I\left(S_{2} | \bm{x}\right)\right], \ldots, \mathbb{E}_{\bm{x} \in \text{test}, c}\left[I\left(S_{d} | \bm{x}\right)\right]\right]^{\top} \in \mathbb{R}^d$ denotes interaction effects of all $m$-order interactive concepts, which are averaged over different testing samples in the category $c$.

In addition, the similarity is defined as the following Jaccard similarity between non-negative elements of $I_{\text{train, c}}^{(m)}$ and $I_{\text{test, c}}^{(m)}$. 

\begin{equation}
    \operatorname{sim}\left(I_{\text{train}, c}^{(m)}, I_{\text{test}, c}^{(m)}\right)=\operatorname{Jaccard} \operatorname{sim}\left(\tilde{I}_{\text{train}, c}^{(m)}, \tilde{I}_{\text{test}, c}^{(m)}\right)=\frac{\left\|\min \left(\tilde{I}_{\text {train }, c}^{(m)}, \tilde{I}_{\text{test}, c}^{(m)}\right)\right\|_1}{\left\|\max \left(\tilde{I}_{\text{train}, c}^{(m)}, \tilde{I}_{\text{test}, c}^{(m)}\right)\right\|_1},
\end{equation}
where the $2d$-dimensional vector $\tilde{I}_{\text{train}, c}^{(m)}=\left[\left(\max \left(I_{\text{train}, c}^{(m)}, 0\right)\right)^{\top},-\left(\min \left(I_{\text{train}, c}^{(m)}, 0\right)\right)^{\top}\right]^\top$ is constructed to contain non-negative elements of ${I}_{\text{train}, c}$. Similarly, $\tilde{I}_{\text{test}, c}$ is constructed based on ${I}_{\text{test}, c}$ to contain non-negative elements. Thus, a high Jaccard similarity indicates that most $m$-order interactive concepts can be well-generalized from training samples to testing samples.

In this experiment, we train 8-layer MLPs for tabular datasets, including the Census dataset and the TV news dataset. All MLPs contain 100 neurons in each hidden layer. For each DNN, we compute the interaction effects of all interactive concepts encoded by the network.
Then, we follow~\citet{zhou2023concept} to evaluate the average generalization ability $g^{(m)}$ of interactive concepts of different complexities (orders) on the above-mentioned DNNs. Table \ref{tab:concept-generalization} shows that complex (high-order) interactive concepts usually have poorer generalization power than simple (low-order) interactive concepts.

\begin{table}[h]
\centering
\caption{Comparison of the generalization ability $g^{(m)}$ of interactive concepts of different orders.}
 \vspace{0.2cm}
\begin{tabular}{lccccc} 
\toprule
& order=1 & order=3 & order=5 & order=7 & order=9 \\ 
\hline
MLP-8 on Census & 0.7989 & 0.6203 & 0.5505 & 0.4436 & 0.3758 \\
MLP-8 on TV news & 0.8156 & 0.5854 & 0.3860 & 0.3322 & 0.1522 \\ 
\bottomrule
\end{tabular}
\label{tab:concept-generalization}
\end{table}

\newpage

\section{Proof of Theorems}\label{apdx:proof-of-theorems}
\subsection{Proof of Lemma \ref{lemma:taylor_interaction} in the main paper}\label{apdx:proof_taylor_interaction} 

\textbf{Lemma \ref{lemma:taylor_interaction}.} 
\textit{Given a neural network $v$ and an arbitrary input sample {\small $\boldsymbol{x}'\in \mathbb{R}^n$}, 
the network output can be decomposed using the Taylor expansion {\small $v(\boldsymbol{x}') = \sum\nolimits_{S \subseteq N} \sum\nolimits_{\boldsymbol{\pi} \in {Q_S}} U_{S, \boldsymbol{\pi}} \cdot J(S,\boldsymbol{\pi}|\boldsymbol{x}')$}.
In this way, according to Eq.~(3) in the main paper, the interaction effect {\small $I(S|\bm{x}')$} on the sample $\bm{x}'$ can be reformulated as
\begin{equation}\label{eqn:apdxrewriteI(S)}
\begin{small}
\begin{aligned}
I(S|\boldsymbol{x}')   =
    \sum\nolimits_{\boldsymbol{\pi} \in Q_S} U_{S, \boldsymbol{\pi}} \cdot J(S,\boldsymbol{\pi}|\boldsymbol{x}'), 
\end{aligned}
\end{small}
\end{equation}
where  {\small $J(S,\boldsymbol{\pi}|\boldsymbol{x}')=\prod\nolimits_{i \in S} \left({\rm sign}(x'_i - r_i) \cdot \frac{x'_i-{r}_{i}}{\tau}\right)^{\pi_{i}}$} denotes an expansion term of the degree $\boldsymbol{\pi}$, {\small $\boldsymbol{\pi} \in Q_S = \{[\pi_1, \dots, \pi_n]| \forall i \in S, \pi_i \in \mathbb{N}^+; \forall i \not\in S, \pi_i =0\}$}.
{\small $U_{S,\boldsymbol{\pi}} {=}$ $ \frac{\tau^m}{\prod_{i=1}^n \pi_i!} \frac{\partial^{m} v(\boldsymbol{x}_{\emptyset})}{\partial x_{1}^{\pi_{1}} \cdots \partial x_{n}^{\pi_{n}}} \cdot \prod_{i \in S}  [{\rm sign}(x'_i- r_i)]^{\pi_i}$}, {\small $m = \sum_{i=1}^n \pi_i$}.}

\begin{proof}
Let us denote the function on the right of Eq.~(\ref{eqn:apdxrewriteI(S)}) by {\small$\tilde I(S|\bm{x}^{\prime})$}, \textit{i.e.},  
\begin{equation}
\tilde I(S|\bm{x}^{\prime}) = \sum\nolimits_{\pi \in Q_S} U_{S,\pi} J(S,\pi|\bm{x}^{\prime})
\end{equation}
We need to prove that for any arbitrary input sample {\small$\forall \bm{x}^{\prime} \in \mathbb{R}^n$}, {\small $\tilde I(S|\bm{x}^{\prime}) = I(S|\bm{x}^{\prime})$}.

Actually, it has been proven in \citet{grabisch1999axiomatic} and \citet{ren2021AOG} that the Harsanyi dividend {\small $I(S|\bm{x}^{\prime})$} is the \textbf{unique} metric satisfying the faithfulness requirement mentioned in the main paper, 
\textit{i.e.}, satisfying 
\begin{equation}\label{eqn:apdxfaithfulness}
\forall  \ T\subseteq N, \ v(\bm{x}^{\prime}_T) = \sum\nolimits_{S \in \Omega, S\subseteq T} I(S|\bm{x}^{\prime}). 
\end{equation}
Thus, as long as we can prove that {\small $\tilde I(S|\bm{x}^{\prime})$} also satisfies the above faithfulness requirement, we can obtain {\small $\tilde I(S|\bm{x}^{\prime}) = I(S|\bm{x}^{\prime})$}.

To this end, we only need to prove {\small $\tilde I(S|\bm{x}^{\prime})$} also satisfies the faithfulness requirement in Eq.~(\ref{eqn:apdxfaithfulness}). 
Specifically, given an input sample {\small$\forall \bm{x}^{\prime} \in \mathbb{R}^n$}, 
let us consider the Taylor expansion of the network output {\small $v(\bm{x}_T)$} of an arbitrarily masked sample 
{\small $\bm{x_T}^{\prime} (T \subseteq N)$}, which is expanded at {\small $\bm{x}_{\emptyset}^{\prime} = [r_1, \dots, r_n]$}. Then, we have
	\begin{equation}\label{eqn:expansionx_T1}
		\begin{small}
			\begin{aligned}
			\forall \ T \subseteq N, \quad	v(\bm{x}'_T)  =  \sum_{\pi_1=0}^{\infty}\sum_{\pi_2=0}^{\infty}\dots\sum_{\pi_n=0}^{\infty} & \frac{1}{\prod_{i=1}^n \pi_i!} \frac{\partial^{m} v(\bm{x}'_{\emptyset})}{\partial x_{1}^{\pi_{1}} \cdots \partial x_{n}^{\pi_{n}}} \cdot \prod_{i=1}^n  [(\bm{x}'_T)_i - r_i] ^{\pi_i},
			\end{aligned}
		\end{small}
	\end{equation}
where {\small $\bm{\pi} \in \{ [\pi_1, \dots, \pi_n]|\forall i \in N, \pi_i \in \mathbb{N}\}$} denotes the degree vector of Taylor expansion terms, and {\small $m = \sum_{i=1}^n \pi_i$}. In addition, $r_i$ denotes the reference value to mask the input variable $x_i$.  

According to the definition of the masked sample {\small $\bm{x}'_T$}, we have that all variables in {\small $T$} keep unchanged and other variables are masked to the reference value. 
That is, {\small $\forall \ i \in T$, $(\bm{x}'_T)_i  = x_i;$ $\forall \ i \not\in T$, $(\bm{x}'_T)_i  = r_i$}. 
Hence, we obtain {\small $\forall i \not\in T, [(\bm{x}'_T)_i - r_i]^{\pi_i} = 0$}. 
Then, among all Taylor expansion terms, only terms corresponding to degrees {\small $\bm{\pi}$} in the set  {\small $P = \{[\pi_1, \dots, \pi_n]| \forall i \in T, \pi_i \in \mathbb{N}; \forall i \not\in T, \pi_i =0\}$} may not be zero. 
Therefore, Eq.~(\ref{eqn:expansionx_T1}) can be re-written as 
	\begin{equation}
		\begin{small}
			\begin{aligned}
			\forall \ T \subseteq N, \quad	 v(\bm{x}'_T)  =   \sum_{\bm{\pi} \in P} \ \frac{1}{\prod_{i=1}^n \pi_i!} \frac{\partial^{m} v(\bm{x}'_{\emptyset})}{\partial x_{1}^{\pi_{1}} \cdots \partial x_{n}^{\pi_{n}}} \cdot \prod_{i \in T} (x'_i - r_i) ^{\pi_i}.
			\end{aligned}
		\end{small}
	\end{equation}
We find that the set $P$ can be divided into multiple disjoint sets as follows, {\small $P = \cup_{S \subseteq T} Q_{S}$}, where {\small $Q_{S}= \{[\pi_1, \dots, \pi_n]| \forall i \in S, \pi_i \in \mathbb{N}^+; \forall i \not\in S, \pi_i =0\}$}. 
Then, we can derive that 
	\begin{equation}\label{eqn:final}
		\begin{small}
			\begin{aligned}
				\forall \ T \subseteq N, \quad v(\bm{x}'_T)  &=   \sum_{S \subseteq T}\sum_{\bm{\pi} \in Q_{S}}  \frac{1}{\prod_{i=1}^n \pi_i!} \frac{\partial^{m} v(\bm{x}'_{\emptyset})}{\partial x_{1}^{\pi_{1}} \cdots \partial x_{n}^{\pi_{n}}} \cdot \prod_{i \in S} (x'_i - r_i) ^{\pi_i} \\
				& =   \sum_{S \subseteq T}\sum_{\bm{\pi} \in Q_{S}}   \underbrace{\frac{\tau^m}{\prod_{i=1}^n \pi_i!} \frac{\partial^{m} v(\bm{x}'_{\emptyset})}{\partial x_{1}^{\pi_{1}} \cdots \partial x_{n}^{\pi_{n}}} \prod_{i\in S} (\delta_i)^{\pi_i}}_{\text{termed } U_{S,\pi}}  \cdot \underbrace{\prod_{i \in S} (\delta_i \frac{x'_i - r_i}{\tau}) ^{\pi_i}}_{\text{termed } J(S,\pi|\bm{x}')},  \\
			\end{aligned}
		\end{small}
	\end{equation}
where {\small $\tau \in \mathbb{R}$} is a pre-defined constant and {\small $\delta_i = {\rm sign}(x_i - r_i) \in \{-1,1\}$} is a sign function and it satisfies {\small $\prod_{i \in S}(\delta_i)^{2\pi_i} = 1$}. 
Then, Eq. (\ref{eqn:final}) can be re-written as
	\begin{equation}
		\begin{small}
			\begin{aligned}
				\forall \ T \subseteq N, \    v(\bm{x}'_T) = \sum_{S \subseteq T}\sum_{\bm{\pi} \in Q_S} U_{S,\bm{\pi}} 
				\cdot J(S, \bm{\pi}|\bm{x}')  = \sum_{S \subseteq T} \tilde I(S|\bm{x}').
			\end{aligned}
		\end{small}
	\end{equation}
Thus,  {\small $\tilde I(S|\bm{x}^{\prime})$} satisfies the faithfulness requirement in Eq.~(\ref{eqn:apdxfaithfulness}) when {\small $\Omega = 2^N$}. 

Therefore, Lemma  1 holds.
\end{proof}

%----------------------------------------
\subsection{Proof of Theorem \ref{theorem:variance_degree1} in the main paper}\label{apdx:proof_variance_degree1} 

\textbf{Theorem \ref{theorem:variance_degree1}.} 
\textit{Let {\small $\boldsymbol{\hat \pi}$} denote the lowest degree of the expansion terms of the interaction effect {\small $I(S|\bm{x}')$}, \textit{i.e.}, {\small $\forall i \in S, \hat \pi_i = 1; \forall i \not\in S, \hat \pi_i= 0$}. 
Let us consider the interaction effect {\small $I(S|\bm{x}')$} only containing the expansion term of the lowest degree, \textit{i.e.}, {\small $I(S|\bm{x}') = U_{S, \bm{\hat \pi}}\cdot J(S, \bm{\hat \pi}|\bm{x}')$}.
In this way, the mean and variance of the interaction effect 
{\small $I(S|\bm{x}'=\boldsymbol{x}+\boldsymbol{\epsilon})$} over different perturbations {\small $\boldsymbol{\epsilon}$} are given as
\begin{equation}
\begin{small}
\begin{aligned}
  \mathbb{E}_{\bm{\epsilon}}[I(S|\bm{x}+\bm{\epsilon})]  &= U_{S,\bm{\hat \pi}}, \\ 
{\rm Var}_{\bm{\epsilon}}[I(S|\bm{x}+\bm{\epsilon})] &= U_{S,\bm{\hat \pi}}^2 ((1+(\sigma/\tau )^2)^{|S|} - 1).
\end{aligned}
\end{small}
\end{equation}}

\begin{proof}
If we only consider Taylor expansion term of the lowest degree, then $I(S|\bm{x}') = U_{S, \hat{\bm{\pi}}}\cdot J(S,\hat{\bm{\pi}}|\bm{x}')$, where $J(S,\hat{\bm{\pi}}|\bm{x}') =
\prod\nolimits_{i \in S} {\rm sign}(x'_i - r_i) \cdot \frac{x'_i-{r}_{i}}{\tau}$.

Let us add a Gaussian perturbation $\bm{\epsilon} \sim \mathcal{N}(\bm{0}, \sigma^2\bm{I})$ to the input sample $\bm{x}$. In this way, we have
\begin{equation}
\begin{aligned}
I(S|\bm{x}+\bm{\epsilon}) &\approx U_{S, \hat{\bm{\pi}}}\cdot J(S,\hat{\bm{\pi}}|\bm{x}+\bm{\epsilon})\\
J(S,\hat{\bm{\pi}}|\bm{x}+\bm{\epsilon}) 
&= \prod_{i \in S} {\rm sign}(x_i+\epsilon_i - r_i) \cdot \frac{x_i+{\epsilon}_i-{r}_{i}}{\tau}\\
&= \prod_{i \in S} \left({\rm sign}(x_i+\epsilon_i - r_i) \cdot \frac{x_i-{r}_{i}}{\tau} + {\rm sign}(x_i+\epsilon_i - r_i) \cdot \frac{\epsilon_i}{\tau} \right)\\
\end{aligned}
\end{equation}

According to the setting of the reference value in Section \ref{sec:sensitivity}, we have $\forall i \in S, x_i - r_i \in \{-\tau, \tau\}$.
Also in Section 2.3, we have assumed that the variance of the perturbation $\bm{\epsilon}$ is small, so that we can ignore the extremely low probability that the perturbation is large such that $|\epsilon_i| \ge \tau$. In this way, we have ${\rm sign}(x_i+\epsilon_i-r_i)={\rm sign}(x_i-r_i)$, and we can obtain
\begin{equation}
\begin{aligned}
J(S,\hat{\bm{\pi}}|\bm{x}+\bm{\epsilon})
&=\prod_{i \in S} \left({\rm sign}(x_i - r_i)\cdot \frac{x_i-{r}_{i}}{\tau} + {\rm sign}(x_i - r_i) \cdot \frac{\epsilon_i}{\tau} \right)\\
&=\prod_{i \in S} \left(1 + {\rm sign}(x_i - r_i) \cdot \frac{\epsilon_i}{\tau} \right)
\end{aligned}
\end{equation}
\begin{equation}
\begin{aligned}
\Rightarrow 
\mathbb{E}_{\bm{\epsilon}}[J(S,\hat{\bm{\pi}}|\bm{x}+\bm{\epsilon})] &= \mathbb{E}_{\bm{\epsilon}}\left[ \prod_{i \in S} \left(1 + {\rm sign}(x_i - r_i) \cdot \frac{\epsilon_i}{\tau} \right) \right]\\
{\rm Var}_{\bm{\epsilon}}[J(S,\hat{\bm{\pi}}|\bm{x}+\bm{\epsilon})] &= {\rm Var}_{\bm{\epsilon}} \left[\prod_{i \in S} \left(1 + {\rm sign}(x_i - r_i) \cdot \frac{\epsilon_i}{\tau} \right) \right] \label{eq:mean_var_J(S,pihat|x+eps)}
\end{aligned}
\end{equation}

Since ${\rm sign}(x_i-r_i) \in \{-1,1\}$, we have $1 + {\rm sign}(x_i - r_i) \cdot \frac{\epsilon_i}{\tau} \sim \mathcal{N}(1,(\sigma/\tau)^2), \forall i \in S$.

\begin{proposition} \label{prop:mean_var_of_product}
If random variables $X_1,X_2,\cdots, X_k$ are independent of each other, then $\mathbb{E}[X_1 X_2 \cdots X_k]=\prod_{i=1}^k \mathbb{E}[X_i]$, and ${\rm Var}[X_1 X_2 \cdots X_k]=\prod_{i=1}^k (\mathbb{E}[X_i]^2 + {\rm Var}[X_i]) - \prod_{i=1}^k \mathbb{E}[X_i]^2$.
\end{proposition}

According to the above proposition, we have 
\begin{equation}
\begin{aligned}
\mathbb{E}_{\bm{\epsilon}}[J(S,\hat{\bm{\pi}}|\bm{x}+\bm{\epsilon})] &= \prod_{i \in S} 1 = 1\\
  {\rm Var}_{\bm{\epsilon}}[J(S,\hat{\bm{\pi}}|\bm{x}+\bm{\epsilon})]  &=\prod_{i \in S}\left(1^2+\left({\sigma}/{\tau}\right)^{2}\right)-\prod_{i\in S} 1^{2} \\
&=\left(1+\left({\sigma}/{\tau}\right)^{2}\right)^{|S|}-1
\end{aligned}
\end{equation}

Therefore, 
\begin{equation}
\begin{aligned}
\mathbb{E}_{\bm{\epsilon}}[I(S|\bm{x}+\bm{\epsilon})] & 
 = \mathbb{E}_{\bm{\epsilon}}[U_{S, \hat{\bm{\pi}}} \cdot J(S,\hat{\bm{\pi}}|\bm{x}+\bm{\epsilon})] = U_{S, \hat{\bm{\pi}}}\\
{\rm Var}_{\bm{\epsilon}}[I(S|\bm{x}+\bm{\epsilon})] &
=  {\rm Var}_{\bm{\epsilon}}[ U_{S, \hat{\bm{\pi}}}\cdot J(S,\hat{\bm{\pi}}|\bm{x}+\bm{\epsilon})]   =  U^2_{S, \hat{\bm{\pi}}} \left(\left(1+\left({\sigma}/{\tau}\right)^{2}\right)^{|S|}-1\right)
\end{aligned}
\end{equation}

\end{proof}

%----------------------------------------
\subsection{Proof of Theorem \ref{theorem:variance_degree_k} in the main paper}\label{apdx:proof_variance_degree_k} 

\textbf{Theorem \ref{theorem:variance_degree_k}.}  
\textit{Let {\small $\bm{\pi} \in Q_S = \{[\pi_1, \dots, \pi_n] | \forall i \in S, \pi_i \in \mathbb{N}^+; \forall i \not\in S, \pi_i =0\}$} denote an arbitrary degree. 
Then, the mean and the variance of {\small $J(S, \bm{\pi}|\boldsymbol{{x}}+\boldsymbol{\epsilon})$} over perturbations {\small $\bm{\epsilon}$} are
\begin{equation}
\begin{small}
\begin{aligned}
  \mathbb{E}_{\bm{\epsilon}}[J(S,\bm{\pi}|\bm{{x}}+\bm{\epsilon})] &= \mathbb{E}_{\bm{\epsilon}}[\prod\nolimits_{i \in S}(1 + \frac{\epsilon_i}{\tau})^{\pi_i}], \\
{\rm Var}_{\bm{\epsilon}}[J(S,\bm{\pi}|\bm{{x}}+\bm{\epsilon})] &= {\rm Var}_{\bm{\epsilon}}[\prod\nolimits_{i \in S}(1 + \frac{\epsilon_i}{\tau})^{\pi_i}]
\end{aligned}
\end{small}
\end{equation}}

\begin{proof}
According to Lemma \ref{lemma:taylor_interaction}, given an arbitrary input sample $\bm{x}'$, we have 
\begin{equation}
    J(S,\bm{\pi}|\bm{x}')=\prod\nolimits_{i \in S} \left({\rm sign}(x'_i - r_i) \cdot \frac{x'_i-{r}_{i}}{\tau}\right)^{\pi_{i}}
\end{equation}
Let us add a Gaussian perturbation $\bm{\epsilon} \sim \mathcal{N}(\bm{0}, \sigma^2\bm{I})$ to the input sample $\bm{x}$. In this way, we have
\begin{equation}
\begin{aligned}
J(S, \bm{\pi}|\bm{x}+\bm{\epsilon}) 
&= \prod_{i \in S} \left({\rm sign}(x_i+\epsilon_i - r_i) \cdot \frac{x_i+{\epsilon}_i-{r}_{i}}{\tau}\right)^{\pi_i}\\
&= \prod_{i \in S} \left({\rm sign}(x_i+\epsilon_i - r_i) \cdot \frac{x_i-{r}_{i}}{\tau} + {\rm sign}(x_i+\epsilon_i - r_i) \cdot \frac{\epsilon_i}{\tau} \right)^{\pi_i}\\
\end{aligned}
\end{equation}

According to the setting of the reference value in Section \ref{sec:sensitivity}, $\forall i \in S, x_i - r_i \in \{-\tau, \tau\}$.
Also, in Section \ref{sec:sensitivity}, we have assumed that the variance of the perturbation $\bm{\epsilon}$ is small, so that we can ignore the extremely low probability that the perturbation is large such that $|\epsilon_i| \ge \tau$. In this way, ${\rm sign}(x_i+\epsilon_i-r_i)={\rm sign}(x_i-r_i)$, and we can obtain
\begin{equation}
\begin{aligned}
J(S, \bm{\pi}|\bm{x}+\bm{\epsilon})
&=\prod_{i \in S} \left({\rm sign}(x_i - r_i)\cdot \frac{x_i-{r}_{i}}{\tau} + {\rm sign}(x_i - r_i) \cdot \frac{\epsilon_i}{\tau} \right)^{\pi_i}\\
&=\prod_{i \in S} \left(1 + {\rm sign}(x_i - r_i) \cdot \frac{\epsilon_i}{\tau} \right)^{\pi_i}
\end{aligned}
\end{equation}
\begin{equation}
\begin{aligned}
\Rightarrow 
\mathbb{E}_{\bm{\epsilon}}[J(S, \bm{\pi}|\bm{x}+\bm{\epsilon})] &= \mathbb{E}_{\bm{\epsilon}}\left[ \prod_{i \in S} \left(1 + {\rm sign}(x_i - r_i) \cdot \frac{\epsilon_i}{\tau} \right)^{\pi_i} \right]\\
{\rm Var}_{\bm{\epsilon}}[J(S, \bm{\pi}|\bm{x}+\bm{\epsilon})] &= {\rm Var}_{\bm{\epsilon}} \left[\prod_{i \in S} \left(1 + {\rm sign}(x_i - r_i) \cdot \frac{\epsilon_i}{\tau} \right)^{\pi_i} \right] 
\label{eq:mean_var_J(S,pihat|x+eps)}
\end{aligned}
\end{equation}

Since $\forall i \in S$, $\epsilon_i$ is independent of each other, according to Proposition \ref{prop:mean_var_of_product} and Eq. (\ref{eq:mean_var_J(S,pihat|x+eps)}), we have 
\begin{equation}
\begin{aligned}
\mathbb{E}_{\bm{\epsilon}}[J(S, \bm{\pi}|\bm{x}+\bm{\epsilon})] &= \prod_{i \in S} \mathbb{E}_{{\epsilon_i}}\left[\left(1 + {\rm sign}(x_i - r_i) \cdot \frac{\epsilon_i}{\tau} \right)^{\pi_i} \right] \\
  {\rm Var}_{\bm{\epsilon}}[J(S, \bm{\pi}|\bm{x}+\bm{\epsilon})]  &=\prod_{i \in S}\mathbb{E}_{{\epsilon_i}}\left[\left(1 + {\rm sign}(x_i - r_i) \cdot \frac{\epsilon_i}{\tau} \right)^{2\pi_i} \right] - 
  \prod_{i\in S} \left( \mathbb{E}_{{\epsilon_i}}\left[\left(1 + {\rm sign}(x_i - r_i) \cdot \frac{\epsilon_i}{\tau} \right)^{\pi_i} \right] \right)^2
\end{aligned}
\end{equation}

Since ${\rm sign}(x_i-r_i) \in \{-1,1\}$, we have $\mathbb{E}_{{\epsilon_i}} \left[ \left( 1 + {\rm sign}(x_i - r_i) \cdot \frac{\epsilon_i}{\tau} \right)^{k} \right] = \mathbb{E}_{{\epsilon_i}} \left[ \left( 1 + \frac{\epsilon_i}{\tau} \right)^{k} \right], \forall k \in \mathbb{N}^+$. Therefore, we obtain

\begin{equation*}
\begin{aligned}
\mathbb{E}_{\bm{\epsilon}}[J(S,\bm{\pi}|\bm{x}+\bm{\epsilon})]  &= \prod_{i\in S} \mathbb{E}_{\epsilon_i}\left[\left(1+\frac{\epsilon_{i}}{\tau}\right)^{\pi_{i}}\right]\\
&= \mathbb{E}_{\bm{\epsilon}} \left[ \prod_{i\in S} \left(1+\frac{\epsilon_{i}}{\tau}\right)^{\pi_{i}}\right]\\
{\rm Var}_{\bm{\epsilon}}[J(S,\bm{\pi}|\bm{x}+\bm{\epsilon})]  &= \prod_{i\in S} \mathbb{E}_{\epsilon_i}\left[\left(1+\frac{\epsilon_{i}}{\tau}\right)^{2 \pi_{i}}\right] - \prod_{i \in S}\left(\mathbb{E}_{\epsilon_i}\left[\left(1+\frac{\epsilon_{i}}{\tau}\right)^{\pi_{i}}\right]\right)^{2}\\
&= {\rm Var}_{\bm{\epsilon}} \left[\prod_{i \in S} \left(1 + \frac{\epsilon_i}{\tau} \right)^{\pi_i} \right].
\end{aligned}
\end{equation*}
\end{proof}

%----------------------------------------
\subsection{Proof of Theorem \ref{theorem:high_large_variance} in the main paper}\label{apdx:proof_high_large_variance}

\textbf{Theorem \ref{theorem:high_large_variance}.}  
\textit{Let {\small $S$ and $S'$} be two interactive concepts, such that {\small $S \subsetneq S'$}. 
Let us consider expansion terms {\small $J(S,\bm{\pi})$} and {\small $J(S',\bm{\pi}')$},
where the term {\small $J(S',\bm{\pi}')$} is extended from the term {\small $J(S,\bm{\pi})$} with {\small $\bm{\pi} \prec \bm{\pi}^\prime$}. I.e., (1) {\small $\forall i \in S', \pi'_i \in \mathbb{N}^+$}; otherwise, {\small $\pi'_i = 0$}. (2) Given {\small $\bm{\pi}'$}, {\small $\forall j \in S, \pi_j = \pi'_j$}; otherwise, {\small $\pi_j = 0$}.
Then, we have 
\begin{equation}
\begin{small}
\begin{aligned}
% --- first inequality starts
\frac{{\rm Var}_{\bm{\epsilon}}[J(S', \bm{\pi}'|\bm{{x}}+\bm{\epsilon})]}
{{\rm Var}_{\bm{\epsilon}}[J(S,\bm{\pi}|\bm{{x}}+\bm{\epsilon})]}  &> 
\prod\nolimits_{i \in S' \setminus S}  \mathbb{E}^2_{\epsilon_i}[(1 + \frac{\epsilon_i}{\tau})^{\pi'_i}], \\
\frac
{\mathbb{E}_{\bm{\epsilon}}[J(S',\bm{\pi}'|\bm{{x}}+\bm{\epsilon})]/
{\rm Var}_{\bm{\epsilon}}[J(S', \bm{\pi}'|\bm{{x}}+\bm{\epsilon})]}
{\mathbb{E}_{\bm{\epsilon}}[J(S,\bm{\pi}|\bm{{x}}+\bm{\epsilon})]/{\rm Var}_{\bm{\epsilon}}[J(S,\bm{\pi}|\bm{{x}}+\bm{\epsilon})]}  &<
\frac{1}{\prod\nolimits_{i \in S' \setminus S}  \mathbb{E}_{\epsilon_i}[(1 + \frac{\epsilon_i}{\tau})^{\pi'_i}]},
\end{aligned}
\end{small}
\end{equation}
and we can also obtain {\small $\mathbb{E}_{\epsilon_i}[(1 + \frac{\epsilon_i}{\tau})^{\pi'_i}] \ge 1$}.}

\begin{proof}
According to Theorem \ref{theorem:variance_degree_k}, we have 
\begin{equation}
\begin{aligned}
{\rm Var}_{\bm{\epsilon}}[J(S',\bm{\pi}'|\bm{x}+\bm{\epsilon})]
&= {\rm Var}_{\bm{\epsilon}} \left[\prod_{i \in S'} \left(1 + \frac{\epsilon_i}{\tau} \right)^{\pi'_i} \right]\\
&= {\rm Var}_{\bm{\epsilon}} \left[\prod_{i \in S} \left(1 + \frac{\epsilon_i}{\tau} \right)^{\pi'_i}  \prod_{i \in S'\setminus S} \left(1 + \frac{\epsilon_i}{\tau} \right)^{\pi'_i} \right] 
\quad \ // S \subsetneq S'\\
&= {\rm Var}_{\bm{\epsilon}} \left[\underbrace{\prod_{i \in S} \left(1 + \frac{\epsilon_i}{\tau} \right)^{\pi_i}}_{A}  \underbrace{\prod_{i \in S'\setminus S} \left(1 + \frac{\epsilon_i}{\tau} \right)^{\pi'_i}}_{B} \right] 
\quad \ // \forall i \in S, \pi'_i=\pi_i\\
&= {\rm Var}_{\bm{\epsilon}}[AB] \\
&= (\mathbb{E}^2_{\bm{\epsilon}}[A] + {\rm Var}_{\bm{\epsilon}}[A])(\mathbb{E}^2_{\bm{\epsilon}}[B]+ {\rm Var}_{\bm{\epsilon}}[B]) - \mathbb{E}^2_{\bm{\epsilon}}[A] \mathbb{E}^2_{\bm{\epsilon}}[B] \\
& \quad \ // \text{$A$ and $B$ are independent; Proposition \ref{prop:mean_var_of_product}}\\
&= \mathbb{E}^2_{\bm{\epsilon}}[A] {\rm Var}_{\bm{\epsilon}}[B] + \mathbb{E}^2_{\bm{\epsilon}}[B] {\rm Var}_{\bm{\epsilon}}[A] +  {\rm Var}_{\bm{\epsilon}}[A]{\rm Var}_{\bm{\epsilon}}[B]\\
&>  \mathbb{E}^2_{\bm{\epsilon}}[B] {\rm Var}_{\bm{\epsilon}}[A] +  {\rm Var}_{\bm{\epsilon}}[A]{\rm Var}_{\bm{\epsilon}}[B]
\end{aligned}
\end{equation}

Therefore, we can prove the first equality as follows.
\begin{equation}
\begin{aligned}
\frac{{\rm Var}_{\bm{\epsilon}}[J(S',\bm{\pi}'|\bm{x}+\bm{\epsilon})]}{{\rm Var}_{\bm{\epsilon}}[J(S,\bm{\pi}|\bm{x}+\bm{\epsilon})]} 
&= \frac{{\rm Var}_{\bm{\epsilon}}[AB]}{{\rm Var}_{\bm{\epsilon}}[A]}\\
&>  \mathbb{E}^2_{\bm{\epsilon}}[B]  + {\rm Var}_{\bm{\epsilon}}[B]\\
&>  \mathbb{E}^2_{\bm{\epsilon}}[B]\\
&= \mathbb{E}^2_{\bm{\epsilon}} \left[\prod_{i \in S'\setminus S} \left(1 + \frac{\epsilon_i}{\tau} \right)^{\pi'_i} \right]\\
&= \prod_{i \in S'\setminus S} \mathbb{E}^2_{{\epsilon}_i} \left[ \left(1 + \frac{\epsilon_i}{\tau} \right)^{\pi'_i} \right]\\
&\quad \ // \text{$\epsilon_i$ is independent of each other; Proposition \ref{prop:mean_var_of_product}}
\end{aligned}
\end{equation}

Furthermore, we have
\begin{equation}
\begin{aligned}
\mathbb{E}_{\bm{\epsilon}}[J(S',\bm{\pi}'|\bm{x}+\bm{\epsilon})]
&= \mathbb{E}_{\bm{\epsilon}} \left[\prod_{i \in S'} \left(1 + \frac{\epsilon_i}{\tau} \right)^{\pi'_i} \right]\\
&= \mathbb{E}_{\bm{\epsilon}} \left[\prod_{i \in S} \left(1 + \frac{\epsilon_i}{\tau} \right)^{\pi'_i}  \prod_{i \in S'\setminus S} \left(1 + \frac{\epsilon_i}{\tau} \right)^{\pi'_i} \right] 
\quad \ // S \subsetneq S'\\
&= \mathbb{E}_{\bm{\epsilon}} \left[{\prod_{i \in S} \left(1 + \frac{\epsilon_i}{\tau} \right)^{\pi_i}}  {\prod_{i \in S'\setminus S} \left(1 + \frac{\epsilon_i}{\tau} \right)^{\pi'_i}} \right] 
\quad \ // \forall i \in S, \pi'_i=\pi_i\\
&= \mathbb{E}_{\bm{\epsilon}}[AB] \\
\end{aligned}    
\end{equation}
and also 
\begin{equation}
\begin{aligned}
\mathbb{E}_{\bm{\epsilon}}[J(S,\bm{\pi}|\bm{x}+\bm{\epsilon})]
= \mathbb{E}_{\bm{\epsilon}} \left[\prod_{i \in S} \left(1 + \frac{\epsilon_i}{\tau} \right)^{\pi_i} \right] = \mathbb{E}_{\bm{\epsilon}}[A].
\end{aligned}    
\end{equation}
Therefore, we have 
\begin{equation}
\begin{aligned}
\frac{\mathbb{E}_{\bm{\epsilon}}[J(S',\bm{\pi}'|\bm{x}+\bm{\epsilon})]}{\mathbb{E}_{\bm{\epsilon}}[J(S,\bm{\pi}|\bm{x}+\bm{\epsilon})]} = \frac{\mathbb{E}_{\bm{\epsilon}}[AB]}{\mathbb{E}_{\bm{\epsilon}}[A]} = \mathbb{E}_{\bm{\epsilon}}[B].
\end{aligned}    
\end{equation}

Then, we can prove the second inequality as follows.
\begin{equation}
\begin{aligned}
&\frac{\mathbb{E}_{\bm{\epsilon}}[J(S',\bm{\pi}'|\bm{{x}}+\bm{\epsilon})]/{\rm Var}_{\bm{\epsilon}}[J(S', \bm{\pi}'|\bm{{x}}+\bm{\epsilon})]}
{\mathbb{E}_{\bm{\epsilon}}[J(S,\bm{\pi}|\bm{{x}}+\bm{\epsilon})]
/{\rm Var}_{\bm{\epsilon}}[J(S,\bm{\pi}|\bm{{x}}+\bm{\epsilon})]} \\
= & \frac{\mathbb{E}_{\bm{\epsilon}}[B]}{ {\rm Var}_{\bm{\epsilon}}[AB] / {\rm Var}_{\bm{\epsilon}}[A]}\\
< & \frac{\mathbb{E}_{\bm{\epsilon}}[B]}{ \mathbb{E}^2_{\bm{\epsilon}}[B]}\\
= & \frac{1}{\mathbb{E}_{\bm{\epsilon}}[B]}\\
= & \frac{1}{\mathbb{E}_{\bm{\epsilon}} \left[\prod_{i \in S'\setminus S} \left(1 + \frac{\epsilon_i}{\tau} \right)^{\pi'_i} \right]}\\
= & \frac{1}{\prod_{i \in S'\setminus S} \mathbb{E}_{{\epsilon}_i} \left[ \left(1 + \frac{\epsilon_i}{\tau} \right)^{\pi'_i} \right]}
\end{aligned}
\end{equation}

Moreover, we can prove that $\mathbb{E}_{\epsilon_i}[(1+\frac{\epsilon_i}{\tau})^k] \ge 1, \forall k\in \mathbb{N}^+$, \emph{i.e.}, $\mathbb{E}[X^{k}] \ge 1$, where $X\sim \mathcal{N}(1,(\sigma/\tau)^2)$.

For a random variable following a Gaussian distribution $\tilde{X} \sim \mathcal{N}(\tilde{\mu}, \tilde{\sigma}^2)$, ~\citet{WILLINK2005271} proved the following property:
\begin{equation}
\mathbb{E}\left[\tilde{X}^{k+1}\right]=\tilde{\mu} \mathbb{E}\left[\tilde{X}^{k}\right]+k \tilde{\sigma}^{2} \mathbb{E}\left[\tilde{X}^{k-1}\right]
\end{equation}
Now let us consider $X\sim \mathcal{N}(1,(\sigma/\tau)^2)$. We have $\mathbb{E}\left[{X}^{k+1}\right]= \mathbb{E}\left[{X}^{k}\right] + k ({\sigma}/\tau)^{2} \mathbb{E}\left[{X}^{k-1}\right]$. By induction, it is easy to prove that $\mathbb{E}[X^{k}]\ge \mathbb{E}[X] = 1$.
\end{proof}

%----------------------------------------
\subsection{Proof of Theorem \ref{theorem:w_relationship} in the main paper}\label{apdx:proof_w_relationship}

\textbf{Theorem \ref{theorem:w_relationship}.} 
\textit{Given two random interactive concepts {\small $S$} and {\small $S'$}, we can roughly assume that  {\small $C_S(\bm{x}+\bm{\epsilon})$} is independent of {\small $C_{S'}(\bm{x}+\bm{\epsilon})$}, because the two concepts {\small$S$} and {\small$S'$} usually have little overlap in most cases. Let {\small$\mathbb{E}_{\bm{\epsilon}} [C_S(\bm{x}+\bm{\epsilon})]$} and {\small${\rm Var}_{\bm{\epsilon}} [C_S(\bm{x}+\bm{\epsilon})]$} denote the mean and the variance of  {\small $C_S(\bm{x}+\bm{\epsilon})$} w.r.t. {\small $\bm{\epsilon}$}, respectively. 
Then, the solution  to Eq. (15) in the main paper satisfies the following property:
\begin{equation}
\begin{small}
\begin{aligned}
      \forall \ S \in \Omega, \quad {|U^*_S|} \propto \vert \mathbb{E}_{\bm{\epsilon}} [C_S(\bm{x}+\bm{\epsilon})] / {\rm Var}_{\bm{\epsilon}} [C_S(\bm{x}+\bm{\epsilon})]\vert
\end{aligned}
\end{small}
\end{equation}}

\begin{proof}
Let $p=|\Omega|$. Let  $\bm{C}(\bm{x}+\bm{\epsilon})=[C_{S_1}(\bm{x}+\bm{\epsilon}),\cdots, C_{S_p}(\bm{x}+\bm{\epsilon})]^\top$ denote the vector of all $C_S(\bm{x}+\bm{\epsilon}), S\in \Omega$, and let $\bm{U}=[U_{S_1},\cdots, U_{S_p}]^\top$ denote the vector of all coefficients $U_S, S\in \Omega$.  
To further simplify the notation, we simply use $\bm{C}$ to denote the random vector $\bm{C}(\bm{x}+\bm{\epsilon})$. Besides, since we assume that each dimension of the vector $\bm{C}(\bm{x}+\bm{\epsilon})$ is independent of each other, we can use $\mathbb{E}_{\bm{\epsilon}}[\bm{C}]=[\alpha_1, \cdots, \alpha_p]^\top \in \mathbb{R}^p$ and ${\rm Var}_{\bm{\epsilon}}[\bm{C}]={\rm diag}(\beta^2_1,\cdots, \beta_p^2) \in \mathbb{R}^{p\times p}$ to denote the mean vector and covariance matrix of the random vector $\bm{C}(\bm{x}+\bm{\epsilon})$, respectively.
We prove this theorem in three steps.

% -------- step1 ----------
\textbf{Step 1. We first prove that the optimal solution to Eq.~(\ref{eqn:MSEloss2}) in the main paper is given by
\begin{equation}
\forall 1\le i \le p, \quad    U^*_{S_i} = \frac{1}{\det \bm{M}} \det (\bm{M}_1, \cdots, \bm{M}_{i-1}, \bm{\rho}, \bm{M}_{i+1}, \cdots, \bm{M}_{p}) \label{eq:determinant_solution}
\end{equation}
where $\bm{M}=\mathbb{E}_{\bm{\epsilon}}[\bm{C}]\mathbb{E}_{\bm{\epsilon}}[\bm{C}]^\top + {\rm Var}_{\bm{\epsilon}}[\bm{C}]$,  $\bm{\rho}={y}^* \mathbb{E}_{\bm{\epsilon}}[\bm{C}]$, and $\bm{M}_j$ denotes the $j$-th column of the matrix $\bm{M}$.}

We can rewrite the objective function in Eq.~(\ref{eqn:MSEloss2}) in the main paper as
\begin{equation}
    \min_{\bm{U}} \mathbb{E}_{\bm{\epsilon}}[(y^*-\bm{U}^{\top} \bm{C}(\bm{x}+\bm{\epsilon}))^{2}]
\end{equation}
To minimize the loss $L=\mathbb{E}_{\bm{\epsilon}}[(y^*-\bm{U}^{\top} \bm{C})^{2}]$, we set the gradient of the loss \emph{w.r.t} $\bm{U}$ to zero, \emph{i.e.}, 
\begin{equation}
    \begin{aligned}
    \nabla_{\bm{U}} L &= \mathbb{E}_{\bm{\epsilon}}[ 2\bm{C} (\bm{U}^{\top}\bm{C} - y^*) ]\\
    &= 2 \mathbb{E}_{\bm{\epsilon}}[ \bm{C}\bm{C}^\top \bm{U} - y^* \bm{C} ]\\
    &= 2 \mathbb{E}_{\bm{\epsilon}}[ \bm{C}\bm{C}^\top] \bm{U} -2 y^* \mathbb{E}_{\bm{\epsilon}}[\bm{C} ]\\
    &= 2 (\mathbb{E}_{\bm{\epsilon}}[ \bm{C}] \mathbb{E}_{\bm{\epsilon}}[\bm{C}]^\top + {\rm Var}_{\bm{\epsilon}}[\bm{C}]) \bm{U} -2 y^* \mathbb{E}_{\bm{\epsilon}}[\bm{C} ] = 0
\end{aligned}
\end{equation}
\begin{equation}
    \Rightarrow  (\mathbb{E}_{\bm{\epsilon}}[\bm{C}] \mathbb{E}_{\bm{\epsilon}}[\bm{C}]^{\top}+{\rm Var}_{\bm{\epsilon}}[\bm{C}]) \bm{U}= y^*\mathbb{E}_{\bm{\epsilon}}[\bm{C}] \label{eq:gradient=zero}
\end{equation}

Let $\bm{M}=\mathbb{E}_{\bm{\epsilon}}[\bm{C}]\mathbb{E}_{\bm{\epsilon}}[\bm{C}]^\top + {\rm Var}_{\bm{\epsilon}}[\bm{C}]$, and  $\bm{\rho}={y}^* \mathbb{E}_{\bm{\epsilon}}[\bm{C}]$. By Cramer's rule, we can obtain the solution to Eq. (\ref{eq:gradient=zero}):
$$\forall 1\le i \le p, \quad U^*_{S_i} = \frac{1}{\det \bm{M}} \det (\bm{M}_1, \cdots, \bm{M}_{i-1}, \bm{\rho}, \bm{M}_{i+1}, \cdots, \bm{M}_{p})$$
where $\bm{M}_j$ denotes the $j$-th column of the matrix $\bm{M}$.

% -------- step2 ----------
\textbf{Step 2. We prove that for the optimal solution $\bm{U}^*$, we have
\begin{equation}
    \forall 1\le i,j \le p, \quad \frac{|U^*_{S_i}|}{|U^*_{S_j}|} = \frac{|\mathbb{E}_{\bm{\epsilon}}[C_{S_i}(\bm{x}+\bm{\epsilon})] / {\rm Var}_{\bm{\epsilon}}[C_{S_i}(\bm{x}+\bm{\epsilon})]|}{|\mathbb{E}_{\bm{\epsilon}}[C_{S_j}(\bm{x}+\bm{\epsilon})] / {\rm Var}_{\bm{\epsilon}}[C_{S_j}(\bm{x}+\bm{\epsilon})]|}
\end{equation}
}

Since $\bm{M}=\mathbb{E}_{\bm{\epsilon}}[\bm{C}]\mathbb{E}_{\bm{\epsilon}}[\bm{C}]^\top + {\rm Var}_{\bm{\epsilon}}[\bm{C}]$, we can obtain the $j$-th column of $\bm{M}$ as 
\begin{equation}
    \bm{M}_j  =  \alpha_j \mathbb{E}_{\bm{\epsilon}}[\bm{C}] + \bm{V}_j \label{eq:jth-column-of-M}
\end{equation}
where $\mathbb{E}_{\bm{\epsilon}}[\bm{C}]=[\alpha_1,\cdots,\alpha_p]^\top$, and $ \bm{V}_j= [0, \cdots, \beta_j^2, \cdots, 0]^\top$.

According to the conclusion in Step 1,  we have
\begin{align}
|U^*_{S_i}| & = |\frac{1}{\det \bm{M}}| \cdot |\det(\bm{M}_1, \cdots, \bm{M}_{i-1}, \bm{\rho}, \bm{M}_{i+1}, \cdots, \bm{M}_{j-1}, \bm{M}_{j}, \bm{M}_{j+1}, \cdots, \bm{M}_p)| \\
|U^*_{S_j}| & = |\frac{1}{\det \bm{M}}| \cdot |\det(\bm{M}_1, \cdots, \bm{M}_{i-1}, \bm{M}_i, \bm{M}_{i+1}, \cdots, \bm{M}_{j-1}, \bm{\rho}, \bm{M}_{j+1}, \cdots, \bm{M}_p)|
\end{align}

We know that exchanging the rows or columns of a matrix only changes the sign of the determinant of the matrix, but does not change the absolute value of the determinant. Therefore, we have
\begin{equation}
\begin{aligned}
|U^*_{S_i}|
& = |\frac{1}{\det \bm{M}}| \cdot |\det(\bm{M}_{j},\bm{\rho}, \bm{M}_1, \cdots, \bm{M}_{i-1}, \bm{M}_{i+1}, \cdots, \bm{M}_{j-1}, \bm{M}_{j+1}, \cdots, \bm{M}_p)|\\
& =  |\frac{1}{\det \bm{M}}| \cdot |\det(\bm{M}_{j},\bm{\rho}, \bm{M}_{\text{others}})| \ \quad  // \text{Let $\bm{M}_{\text{others}}$ denote the third to the last column} \\
& =  |\frac{1}{\det \bm{M}}| \cdot |\det(\alpha_j \mathbb{E}_{\bm{\epsilon}}[\bm{C}] + \bm{V}_j, y^* \mathbb{E}_{\bm{\epsilon}}[\bm{C}], \bm{M}_{\text{others}})| \ \quad  // \text{Eq. (\ref{eq:jth-column-of-M}) } \\
& = |\frac{1}{\det \bm{M}}| \cdot |\underbrace{\det(\alpha_j \mathbb{E}_{\bm{\epsilon}}[\bm{C}] , y^* \mathbb{E}_{\bm{\epsilon}}[\bm{C}], \bm{M}_{\text{others}})}_{=0} + \det(\bm{V}_j,y^* \mathbb{E}_{\bm{\epsilon}}[\bm{C}] , \bm{M}_{\text{others}})|\\
&\ \quad  // \text{The determinant is 0 if two columns are linearly dependent}\\
&= |\frac{1}{\det \bm{M}}| \cdot |\det( \bm{V}_j , y^* \mathbb{E}_{\bm{\epsilon}}[\bm{C}], \bm{M}_{\text{others}})|\\
&= |\frac{1}{\det \bm{M}}| \cdot |\det
\begin{bmatrix}
0 & y^* \alpha_1 & \alpha_1\alpha_1+\beta_1^2  & \cdots & \alpha_1 \alpha_p  \\
\vdots \\
0   & y^* \alpha_i &  \alpha_i \alpha_1  & \cdots & \alpha_i\alpha_p \\
\vdots \\
\beta_j^2 & y^* \alpha_j &  \alpha_j \alpha_1 & \cdots & \alpha_j \alpha_p \\
\vdots \\
0 & y^* \alpha_p & \alpha_p\alpha_1  & \cdots & \alpha_p \alpha_p +  \beta_p^2 \\ 
\end{bmatrix} |\\ % end of matrix
& =  |\frac{1}{\det \bm{M}}| \cdot | det
\begin{bmatrix}
\beta_j^2 & y^* \alpha_j &  \alpha_j \alpha_1 & \cdots & \alpha_j \alpha_p \\
0   & y^* \alpha_i &  \alpha_i \alpha_1  & \cdots & \alpha_i\alpha_p \\
0 & y^* \alpha_1 & \alpha_1\alpha_1+\beta_1^2  & \cdots & \alpha_1 \alpha_p  \\
\vdots \\
0 & y^* \alpha_p & \alpha_p\alpha_1  & \cdots & \alpha_p \alpha_p +  \beta_p^2 \\
\end{bmatrix} |
\ \quad  // \text{Exchange rows} \\ % end of matrix
&= |\frac{\alpha_i}{\det \bm{M}}| \cdot |\det
\begin{bmatrix}
\beta_j^2 & y^* \alpha_j &  \alpha_j \alpha_1 & \cdots & \alpha_j \alpha_p \\
0   & y^*  &   \alpha_1  & \cdots & \alpha_p \\
0 & y^* \alpha_1 & \alpha_1\alpha_1+\beta_1^2  & \cdots & \alpha_1 \alpha_p  \\
\vdots \\
0 & y^* \alpha_p & \alpha_p\alpha_1  & \cdots & \alpha_p \alpha_p +  \beta_p^2 \\
\end{bmatrix} |
\ \quad  // \text{Extract out $\alpha_i$} \\% end of matrix
& = |\frac{\alpha_i \beta_j^2}{\det \bm{M}}| \cdot |\det \bm{M}'|,
\end{aligned}
\end{equation}
where
\begin{equation}
    \bm{M}' = 
\begin{bmatrix}
y^*  & \alpha_1  & \cdots & \alpha_p \\
y^* \alpha_1 & \alpha_1\alpha_1+\beta_1^2 & \cdots & \alpha_1 \alpha_p  \\
\cdots \\
y^* \alpha_p & \alpha_p\alpha_1  & \cdots & \alpha_p \alpha_p + \beta_p^2 \\
\end{bmatrix} .
\end{equation}

Similarly, we can prove that
\begin{equation}
\begin{aligned}
|U^*_{S_j}|
& = |\frac{\alpha_j \beta_i^2}{\det \bm{M}}| \cdot |\det \bm{M}'|.
\end{aligned}
\end{equation}

Therefore, we have
\begin{equation*}
    \forall 1\le i,j \le p, \quad \frac{|U^*_{S_i}|}{|U^*_{S_j}|} 
    = \frac{|\alpha_i / \beta_i^2|}{|\alpha_j / \beta_j^2|}
    = \frac{|\mathbb{E}_{\bm{\epsilon}}[C_{S_i}(\bm{x}+\bm{\epsilon})] / {\rm Var}_{\bm{\epsilon}}[C_{S_i}(\bm{x}+\bm{\epsilon})]|}{|\mathbb{E}_{\bm{\epsilon}}[C_{S_j}(\bm{x}+\bm{\epsilon})] / {\rm Var}_{\bm{\epsilon}}[C_{S_j}(\bm{x}+\bm{\epsilon})]|}.
\end{equation*}

% -------- step3 ----------
\textbf{Step 3. Based on Step 2, we can directly prove that for the optimal solution $\bm{U}^*$, we have
\begin{equation}
    \forall S\in \Omega, \quad {|U^*_S|} \propto {|\mathbb{E}_{\bm{\epsilon}}[C_{S}(\bm{x}+\bm{\epsilon})] / {\rm Var}_{\bm{\epsilon}}[C_{S}(\bm{x}+\bm{\epsilon})]|}
\end{equation}
}

\end{proof}

%----------------------------------------
\subsection{Proof of Theorem \ref{theorem:C_Supperbound} in the main paper}\label{apdx:proof_C_Supperbound}

\textbf{Theorem \ref{theorem:C_Supperbound}.} 
\textit{Let {\small $A^{\text{min}} = \min_S |U_S|$}
and {\small $A^{\text{max}} = \max_S |U_S|$} denote the lower bound and the upper bound of {\small $\vert U_S \vert$} over all interactive concepts {\small $S$}. 
Then,  for any {\small $S \subseteq N$}, we have
\begin{equation}\label{eqn:apdx_C_Supperbound}
\begin{small}
\begin{aligned}
  A^{\text{min}} \cdot  \frac{ \vert \mathbb{E}_{\bm{\epsilon}} [I(S|\bm{x}+\bm{\epsilon})]\vert} {{\rm Var}_{\bm{\epsilon}} [I(S |\bm{x}+\bm{\epsilon})]} 
  \leq 
  \frac{\vert \mathbb{E}_{\bm{\epsilon}} [C_S(\bm{x}+\bm{\epsilon})]\vert} {{\rm Var}_{\bm{\epsilon}} [C_S(\bm{x}+\bm{\epsilon})]} \leq  
  A^{\text{max}} \cdot  \frac{ \vert \mathbb{E}_{\bm{\epsilon}} [I(S|\bm{x}+\bm{\epsilon})]\vert} {{\rm Var}_{\bm{\epsilon}} [I(S |\bm{x}+\bm{\epsilon})]}
\end{aligned}
\end{small}
\end{equation}}

\begin{proof}
According to Eq.~(\ref{eqn:def_US}) in the main paper, we have {\small $I(S|\bm{x}') = U_S \cdot C_S(\bm{x}')$}. Hence, 
we have 
\begin{equation}\nonumber
\vert \mathbb{E}_{\bm{\epsilon}}[I(S|\bm{x}+\bm{\epsilon})]\vert    = \vert U_S \vert \cdot \vert \mathbb{E}_{\bm{\epsilon}}[C_S(\bm{x}+\bm{\epsilon})]\vert,   \quad 
{\rm Var}_{\bm{\epsilon}}[I(S|\bm{x}+\bm{\epsilon})]  = U_S^2  \cdot  {\rm Var}_{\bm{\epsilon}}[C_S(\bm{x}+\bm{\epsilon})],  
\end{equation}
Therefore, 
\begin{equation}\nonumber
\frac{\vert \mathbb{E}_{\bm{\epsilon}} [C_S(\bm{x}+\bm{\epsilon})]\vert} {{\rm Var}_{\bm{\epsilon}} [C_S(\bm{x}+\bm{\epsilon})]} = \vert U_S \vert \cdot \frac{ \vert \mathbb{E}_{\bm{\epsilon}} [I(S|\bm{x}+\bm{\epsilon})]\vert} {{\rm Var}_{\bm{\epsilon}} [I(S |\bm{x}+\bm{\epsilon})]}
\end{equation}
Then, let {\small $A_S^{\text{min}} = \min_S |U_S|$}
and {\small $A_S^{\text{max}} = \max_S |U_S|$} denote the lower bound and the upper bound of the absolute value {\small $\vert U_S \vert$} over all interactive concepts {\small $S$}, we have 
\begin{equation}\nonumber 
\begin{aligned}
  A_S^{\text{min}} \cdot  \frac{ \vert \mathbb{E}_{\bm{\epsilon}} [I(S|\bm{x}+\bm{\epsilon})]\vert} {{\rm Var}_{\bm{\epsilon}} [I(S |\bm{x}+\bm{\epsilon})]} \leq \frac{\vert \mathbb{E}_{\bm{\epsilon}} [C_S(\bm{x}+\bm{\epsilon})]\vert} {{\rm Var}_{\bm{\epsilon}} [C_S(\bm{x}+\bm{\epsilon})]} \leq  A_S^{\text{max}} \cdot  \frac{ \vert \mathbb{E}_{\bm{\epsilon}} [I(S|\bm{x}+\bm{\epsilon})]\vert} {{\rm Var}_{\bm{\epsilon}} [I(S |\bm{x}+\bm{\epsilon})]}
\end{aligned}
\end{equation}

\end{proof}

%===================================
\section{Experimental details}\label{sec:apdx_exp_detail}

\textbf{Training settings.} 
We trained standard DNNs and BNNs with the same architectures on two image datasets and two tabular datasets.
For image datasets, we trained standard DNNs and BNNs with two architectures. On the MNIST dataset, we trained a standard DNN and a BNN with the 5-layer MLP architecture. On the CIFAR-10 dataset, we trained a standard DNN and a BNN with the LeNet architecture.
On the two tabular datasets, including the UCI TV news dataset (termed \textit{TV news}) and the UCI census income  dataset (termed \textit{census}), we trained standard DNNs and BNNs with the 8-layer MLP architecture.
All MLPs contained 100 neurons in each hidden layer.
For the training of BNNs, the prior distribution of network weights was set to $\mathcal{N}(\bm{W};\bm{0},\bm{I})$, and the number of Monte Carlo sampling of network weights was set to 1.
All standard DNNs and BNNs were trained using the Adam optimizer ~\citep{kingma2015adam} with learning rate 0.001. The 5-layer MLPs (standard DNN and BNN) on the MNIST dataset was trained for 50 epochs. The LeNet (standard DNN and BNN) on the CIFAR-10 dataset was trained for 300 epochs. The 8-layer MLPs (standard DNN and BNN) on tabular datasets were trained for 200 epochs.

\textbf{Implementation details for the calculation of $I(S)$.} Since the computational cost of $I(S)$ was intolerable for image datasets, we applied a sampling-based approximation method to calculate $U_S$. For the CIFAR-10 dataset ($32\times 32$ pixels on each image), we uniformly split each input image into $8\times 8$ patches. Furthermore, we random sampled 12 patches from the central $6\times 6$ region (\emph{i.e.}, we did not sample patches that were on the edges of an image), and considered these patches as input variables for each image. The remaining 52 patches were set to the reference value. Similarly, for the MNIST dataset ($28\times 28$ pixels on each image), we uniformly split each input image into $7\times 7$ patches, and randomly sampled 12 patches from the central $5\times 5$ region.

\textbf{Implementation details of the reference value.} Let {\small $\mathbb{E}_{\bm{x}}[x_i]$} denote the mean value of the $i$-th input dimension over all input samples in the dataset. Then, given an input sample $\bm{x}$, the reference value is set as follows. 
\begin{equation*}
        r_{i}=\left\{\begin{array}{ll}
x_{i}-\tau, & x_{i}>\mathbb{E}_{\bm{x}}[x_i] \\
x_{i}+\tau, & x_{i}<\mathbb{E}_{\bm{x}}[x_i]
\end{array}\right. 
\end{equation*}
where {\small $\tau \in \mathbb{R}$} is a constant.  We set $\tau=0.5$ on all datasets (including the TV news dataset, the Census dataset, the MNIST dataset, and the CIFAR-10 dataset). In our experiments, we assume that input samples have been normalized as follows. First, we subtract the mean value of each input dimension over the whole dataset from the input sample. Second, we divide each dimension of the input sample by the standard deviation of this input dimension over the whole dataset. In this way, input samples
have zero mean and unit variance on each dimension over the whole dataset, \emph{i.e.}, $\forall i \in N, \mathbb{E}_{\bm{x}}[x_i]=0, {\rm Var}_{\bm{x}}[x_i]=1$.

\textbf{Implementation details of the experiment in Section \ref{sec:approximation} of the main paper.}
In Section \ref{sec:approximation} of the main paper, we minimized the KL divergence between the feature distribution in the surrogate DNN model and the feature distribution in the BNN.
The feature distributions in the surrogate DNN model and in the BNN were not Gaussian distributions. Therefore, the KL divergence between the feature distributions did not have a close-form formula. To facilitate the optimization, we simply used two Gaussian distributions to approximate the feature distributions in the surrogate DNN model and in the BNN, and optimized the KL divergence between the two Gaussian distributions.
Besides, we did not consider the dependency between different feature dimensions to simplify the computation.

%===================================
\section{More visualization results for experiments in Section \ref{sec:approximation} of the main paper}\label{sec:apdx_visual_result}

In this subsection, we provided more visualization results to show that the feature distribution of the surrogate DNN model could well approximate the feature distribution of the BNN.

\begin{figure}[t]\centering
\includegraphics[ width=0.96\textwidth]{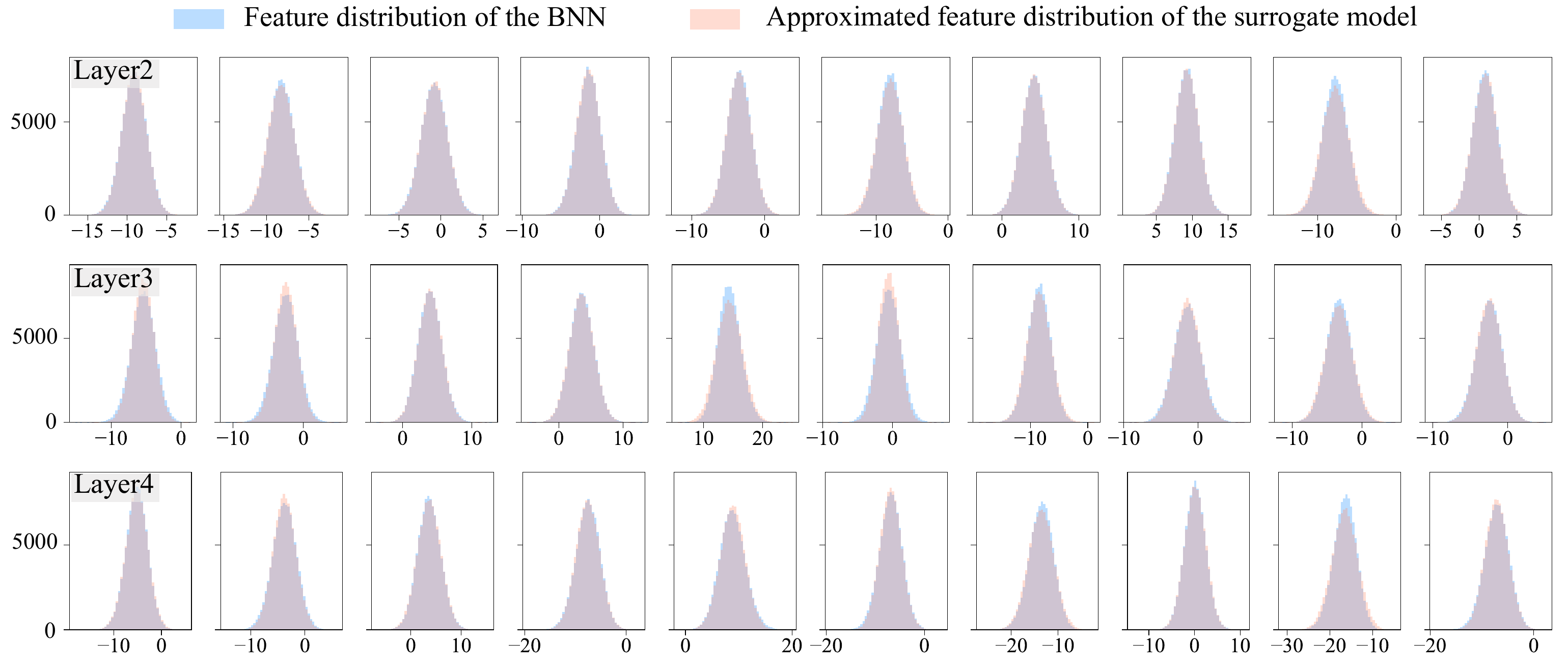}
\caption{More visualization results of MLP-5 on the MNIST dataset.}
\end{figure}

\begin{figure}[t]\centering
\includegraphics[ width=0.96\textwidth]{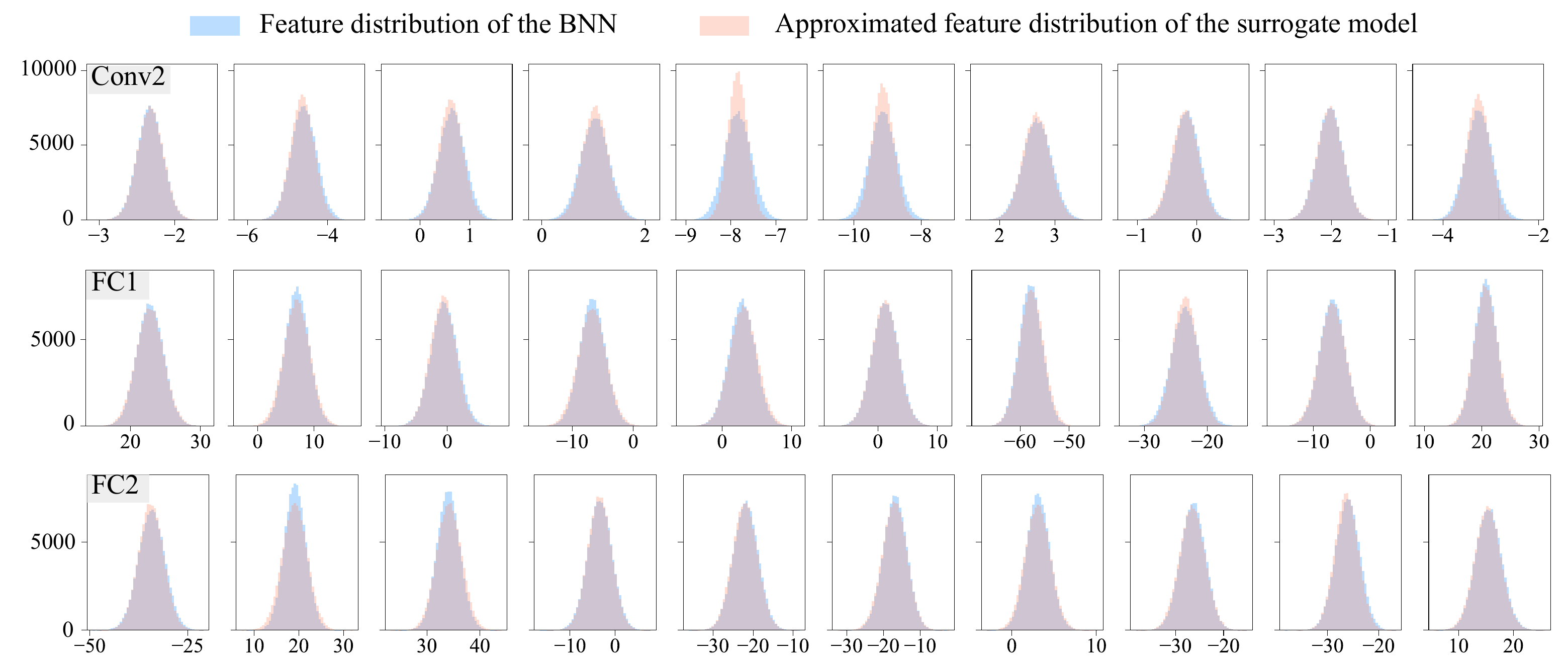}
\caption{More visualization results of LeNet on the CIFAR-10 dataset.}
\end{figure}

\begin{figure}[t]\centering
\includegraphics[ width=0.96\textwidth]{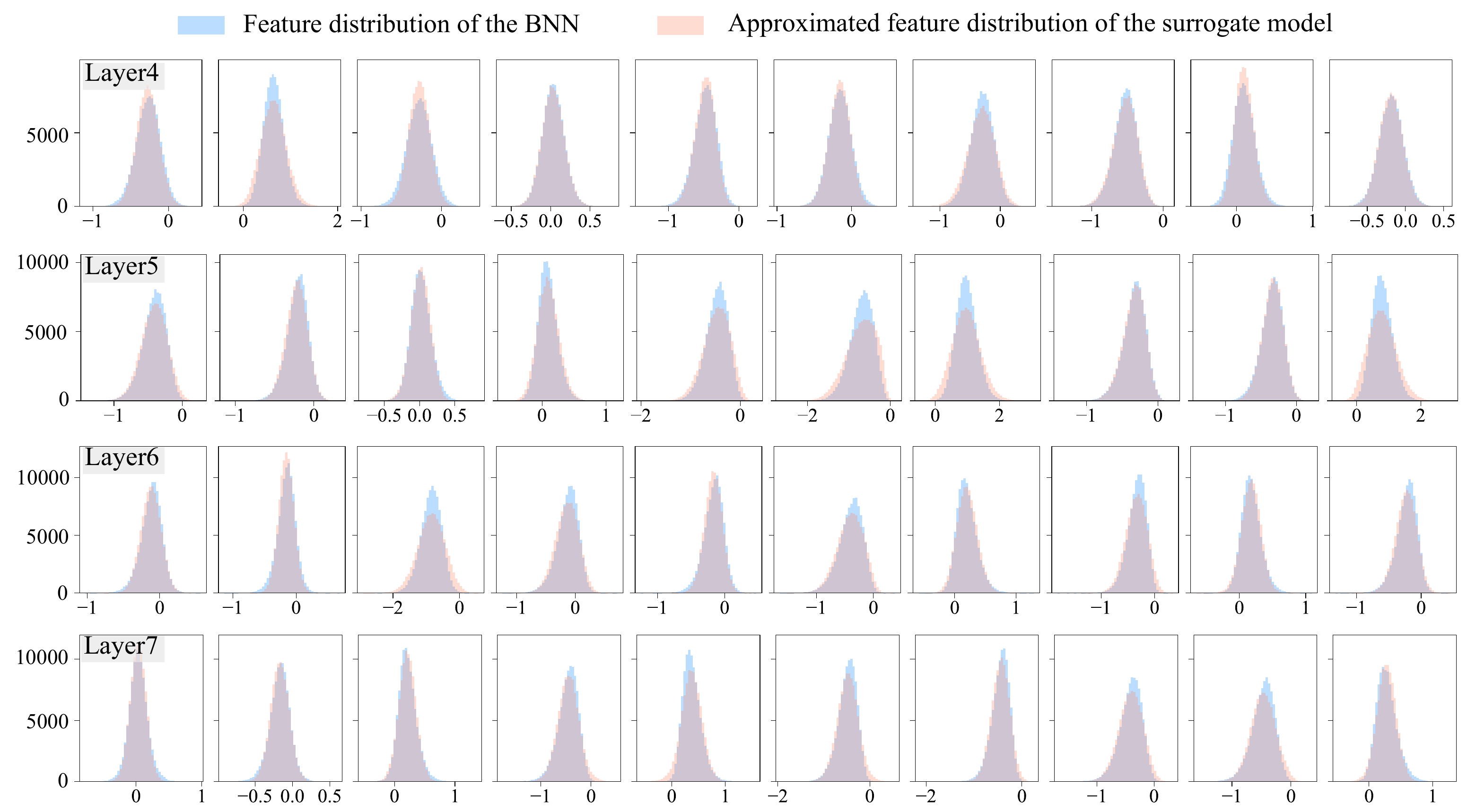}
\caption{More visualization results of MLP-8 on the Census dataset.}
\end{figure}

\begin{figure}[t]\centering
\includegraphics[ width=0.96\textwidth]{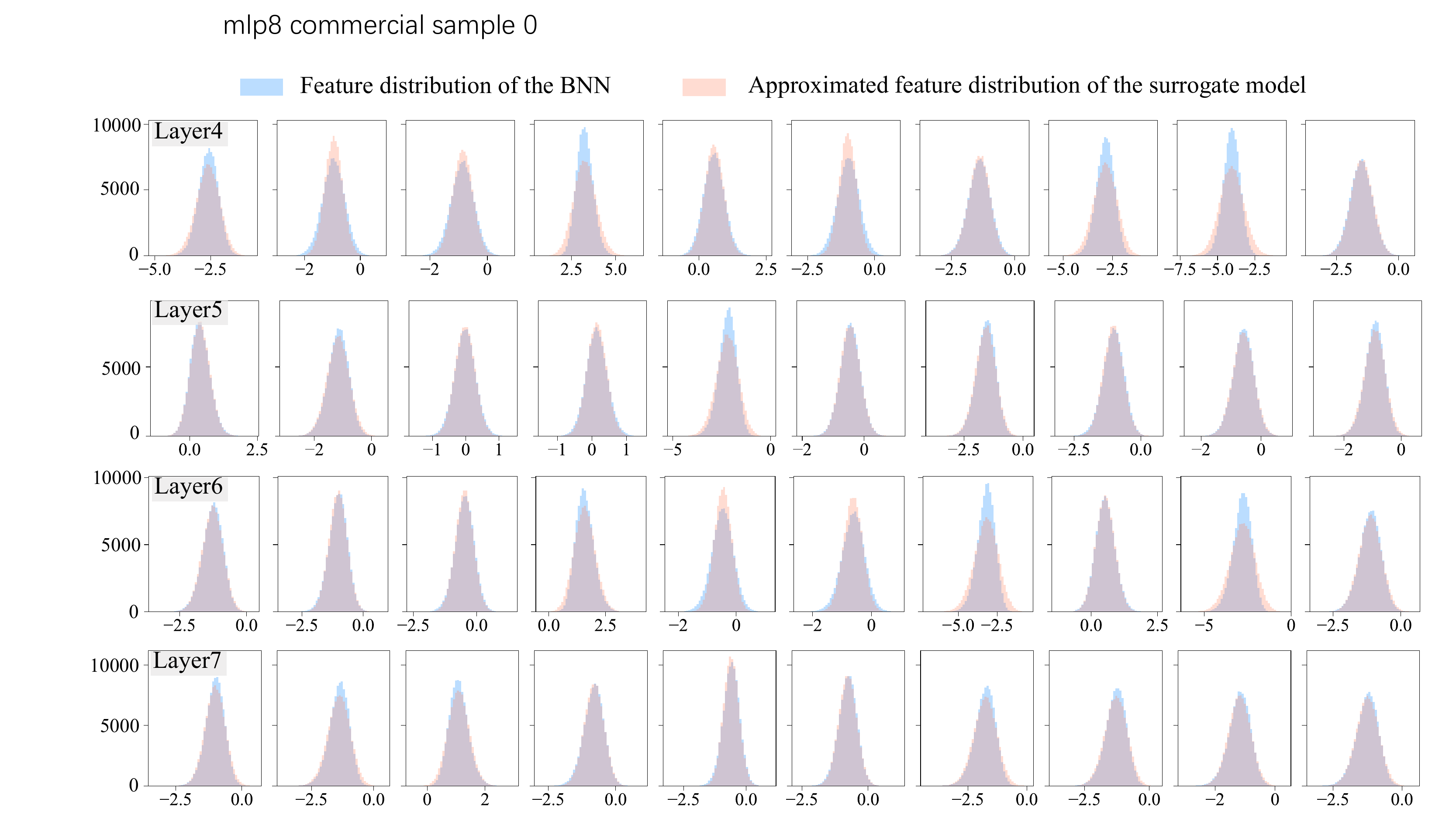}
\caption{More visualization results of MLP-8 on the TV news dataset.}
\end{figure}

%===================================

%%%%%%%%%%%%%%%%%%%%%%%%%%%%%%%%%%%%%%%%%%%%%%%%%%%%%%%%%%%%%%%%%%%%%%%%%%%%%%%
%%%%%%%%%%%%%%%%%%%%%%%%%%%%%%%%%%%%%%%%%%%%%%%%%%%%%%%%%%%%%%%%%%%%%%%%%%%%%%%

\end{document}